\documentclass[10pt,twocolumn,letterpaper]{article}

\usepackage[pagenumbers]{cvpr} 
\usepackage{booktabs}
\usepackage{multirow}
\usepackage{graphicx}
\usepackage{adjustbox}
\usepackage{tabularx} 
\usepackage{cuted}
\usepackage{float}  

\usepackage[absolute]{textpos}
\definecolor{cvprblue}{rgb}{0.21,0.49,0.74}
\usepackage[pagebackref,breaklinks,colorlinks,allcolors=cvprblue]{hyperref}
\usepackage{bbding}
\usepackage{amssymb}
\usepackage{pifont}

\def\paperID{9126} 
\def\confName{CVPR}
\def\confYear{2026}

\title{StreamEQA: Towards Streaming Video Understanding \\
for Embodied Scenarios}

\author{
  Yifei Wang\textsuperscript{1}, 
  Zhenkai Li\textsuperscript{1}, 
  Tianwen Qian\textsuperscript{1\dag}, 
  Huanran Zheng\textsuperscript{1}, 
  Zheng Wang\textsuperscript{2}, \\
  Yuqian Fu\textsuperscript{3}, 
  Xiaoling Wang\textsuperscript{1\dag}\\
  \textsuperscript{1}School of Computer Science and Technology, East China Normal University, \\
  \textsuperscript{2}Zhejiang University of Technology, 
  \textsuperscript{3}INSAIT, Sofia University ``St. Kliment Ohridski'' \\
}

\begin{document}
\maketitle

\begin{textblock*}{9cm}(2.05cm, 2.8cm)  
  \includegraphics[width=0.4\textwidth]{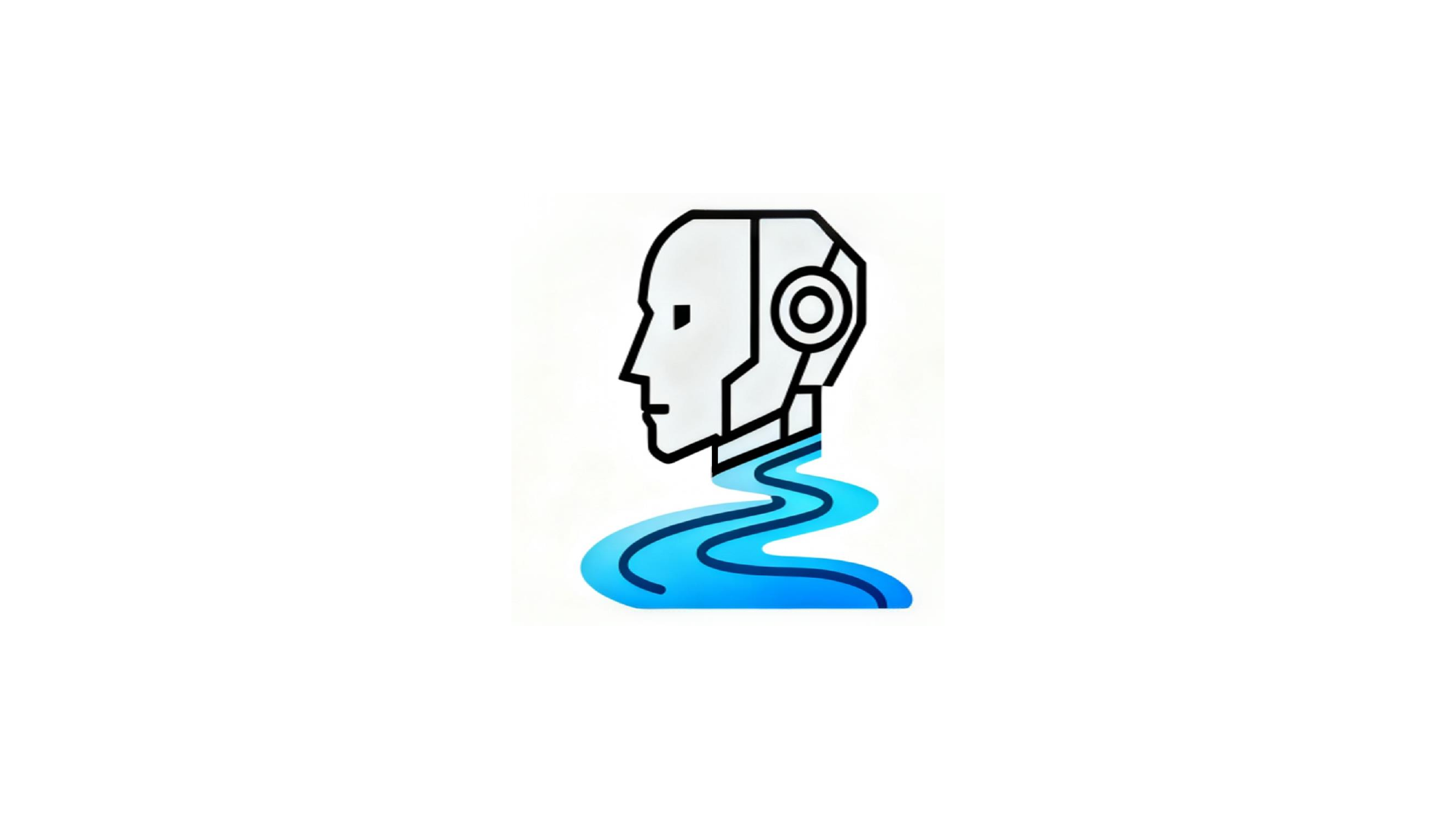}  
  \label{fig:absolute}
\end{textblock*}

\footnotetext[2]{Corresponding authors.}

\begin{abstract}

As embodied intelligence advances toward real-world deployment, the ability to continuously perceive and reason over streaming visual inputs becomes essential. In such settings, an agent must maintain situational awareness of its environment, comprehend the interactions with surrounding entities, and dynamically plan actions informed by past observations, current contexts, and anticipated future events.
To facilitate progress in this direction, we introduce \textbf{StreamEQA}, the first benchmark designed for streaming video question answering in embodied scenarios. StreamEQA evaluates existing MLLMs along two orthogonal dimensions: \textbf{Embodied} and \textbf{Streaming}. Along the embodied dimension, we categorize the questions into three levels: perception, interaction, and planning, which progressively assess a model’s ability to recognize fine-grained visual details, reason about agent–object interactions, and perform high-level goal-directed reasoning. For the streaming dimension, questions are divided into backward, real-time, and forward reasoning, with each mode relying on a distinct temporal context.
Built upon 156 independent long videos, StreamEQA defines 42 tasks and generates approximately 21K question–answer pairs with precise timestamps through a hybrid pipeline combining automated generation and human refinement. Evaluations of 13 state-of-the-art video-LLMs reveal that, despite strong performance on conventional benchmarks, these models still struggle with streaming video understanding in embodied scenarios. We hope StreamEQA will catalyze research on streaming video understanding for embodied applications. Codes and data will be released at \url{https://github.com/MrYF-Wang/StreamEQA}.
\end{abstract}
\section{Introduction}
Unlike traditional AI, Embodied AI~\cite{duan2022survey,li2025clivis,long2025survey} emphasizes dynamic interaction between an agent and its surrounding environment, achieving perception, cognition, and action within a closed feedback loop. Owing to its rich semantics, portability, and low deployment cost, egocentric videos captured by 2D cameras have become the primary sensory channel for embodied agents to perceive the external world. Consequently, video large language models (Video-LLMs)~\cite{tang2025video} are emerging as the core component that enables agents to understand and reason about their surroundings.

As embodied intelligence moves toward real-world deployment, the requirements for video understanding systems have become substantially more demanding. Agents must not only be endowed with fine-grained visual perception, but also extend their capabilities to physical-awared agent-object interactions and high-level task planning. These competencies are essential for complex reasoning and decision-making in embodied scenarios. 
For example, consider an intelligent assistant operating in a kitchen scenario. It must be able to recognize ingredients on the countertop, assess the current stage of recipe preparation, and plan the next cooking steps in alignment with the human’s intention. Meanwhile, sensory input of the agent arrives as continuous video streams, requiring models to maintain ongoing response under contextually limited visual input.

However, existing Video-LLMs~\cite{comanici2025gemini, bai2025qwen2, Zhang2023VideoLLaMAAI} are primarily trained on large-scale internet video–text pairs rather than designed for embodied agent, leading to two fundamental limitations. First, current Video-LLMs excel at perceptual tasks such as action recognition or scene description, but fall short in interaction and planning tasks. Second, mainstream models only support offline video processing, where the full video sequence is accessible during inference. This setup fundamentally differs from the streaming constraint in embodied environments, where the model can only access current and past observations. Such limitations currently prevent these Video-LLMs from being directly deployed on embodied agents to follow human instructions in real-world environments.

To bridge the aforementioned gap, we introduce \textbf{StreamEQA}, a benchmark specifically designed for streaming video question answering in embodied scenarios. Fig.~\ref{fig:fig1} presents representative examples from StreamEQA, along with the key distinctions between our work and existing benchmarks.
The benchmark is structured along two core dimensions, \emph{Embodied} and \emph{Streaming}, to comprehensively evaluate potential of Video-LLMs for embodied deployment. Along the embodied dimension, StreamEQA organizes tasks into three levels: (1) \emph{Perception}, which assesses the ability to recognize objects, their attributes, spatial relationships within the scene; (2) \emph{Interaction}, which measures the understanding of agent–object interactions and their underlying motivations; and (3) \emph{Planning}, which examines a model’s capacity to reflect upon executed actions, adjust current behaviors, and optimize future plans. This hierarchical design enables progressive evaluation of models, from low-level visual recognition to high-level embodied reasoning. Along the streaming dimension, questions are divided into \emph{Backward}, \emph{Real-time}, and \emph{Forward} reasoning modes, each requiring distinct temporal contexts. Specifically, backward reasoning involves retrospection based on past events, real-time reasoning focuses on immediate perceptual interpretation, and forward reasoning demands the anticipation of future outcomes.
Considering the diversity of real-world activities and the availability of existing data, StreamEQA focuses on kitchen scenarios and build upon the HD-EPIC~\cite{Perrett2025HDEPICAH} dataset. By leveraging the densely annotated visual metadata of HD-EPIC, we generate question–answer pairs through a hybrid pipeline that combines automated synthesis with human refinement, ensuring both scale and quality. In total, StreamEQA encompasses 156 unique long videos across 42 task types, yielding approximately 21K QA pairs with precise timestamps.

Based on StreamEQA, we conduct a systematic evaluation of 13 representative Video-LLMs, including proprietary models (\emph{e.g.}, GPT-5~\cite{openai2025gpt5}), open-source models (\emph{e.g.}, Qwen3-VL~\cite{bai2025qwen3}), streaming models (\emph{e.g.}, TimeChat-Online~\cite{Yao2025TimeChatOnline8V}), and egocentric models (\emph{e.g.}, EgoGPT~\cite{yang2025egolife}). Experimental results reveal a significant performance degradation (1.6×↓) compared to the general-purpose benchmark OVO-Bench~\cite{Li2025OVOBenchHF}, highlighting the intrinsic difficulty of streaming video reasoning in embodied scenarios. Moreover, performance analysis shows that interaction and planning tasks are considerably more challenging than basic perception and that the streaming constraint substantially impacts model performance, emphasizing the necessity of developing temporally grounded reasoning mechanisms. 

Our main contributions are summarized as follows:
\begin{itemize}
    \item We propose StreamEQA, the first benchmark for streaming VideoQA in embodied scenarios, featuring both embodied and streaming characteristics, and containing around 21K high-quality, time-stamped QA pairs.
    \item We perform a comprehensive evaluation of 13 state-of-the-art Video-LLMs, quantitatively revealing their limitations in streaming embodied understanding and providing new insights for future research of Video-LLMs.
\end{itemize}

\section{Related Work}
\label{sec:related}
\textbf{Benchmarks for Video Understanding.} Early video QA datasets such as MSVD-QA\cite{10.1145/3123266.3123427}, and MSRVTT-QA~\cite{10.1145/3123266.3123427} focus on short clips, while later benchmarks like TemporalBench~\cite{Cai2024TemporalBenchBF}, and AutoEvalVideo~\cite{Chen2023AutoEvalVideoAA} target causal and temporal reasoning. More comprehensive long-duration evaluation is provided by EgoSchema~\cite{Mangalam2023EgoSchemaAD} and Video-MME~\cite{Fu2024VideoMMETF}, which extend testing to multi-minute or hour-level videos. A variety of benchmarks have been introduced to evaluate video understanding, but most assume offline access to entire videos with the description of explicit video content.

Egocentric and embodied benchmarks~\cite{Grauman2021Ego4D, Ramakrishnan2021HabitatMatterport3D, Dai2017ScanNetR3} primarily examine episodic memory and spatial reasoning. EgoTextVQA~\cite{Zhou2025EgoTextVQATE} targets object-level and text-centric perception, while Nuscenes-QA~\cite{qian2024nuscenes} extends spatial reasoning to driving scenarios.
EgoLife~\cite{yang2025egolife} introduces ultra-long videos, EgoCross~\cite{li2025egocross} and EgoNight~\cite{zhang2025egonight} further challenge models with cross-domain and low-light conditions.
Recent efforts such as EgoThink~\cite{cheng2024egothink} and ECBench~\cite{dang2025ecbench} move toward holistic embodied cognition. Meanwhile, some streaming-oriented datasets, such as MovieChat-1k~\cite{Song2023MovieChatFD}, OVO-Bench~\cite{Li2025OVOBenchHF} and OVBench~\cite{Huang2024OnlineVU}, attempt to model time-aware inference but rarely consider embodied scenarios. However, most existing embodied benchmarks currently conduct evaluations in an offline manner, while streaming benchmarks have limited attention paid to embodied scenarios. In contrast, our work fills this gap by emphasizing the importance of planning targets in real-world and introducing the first comprehensive multi-embodied level testbed for StreamQA.

\noindent
\textbf{MLLMs for Stream Video Understanding.} With the rapid progress of multimodal large language models, video understanding~\cite{Wang2024Qwen2VLEV, fu2025objectrelator, zhou2025camsam2, tan2025xtrack, lin2025neighborretr} has become an increasingly prominent direction of research. Most existing video MLLMs follow an offline paradigm in which sampled frames from an entire video are encoded, commonly via CLIP-based~\cite{Radford2021LearningTV} vision encoders, and projected into the LLM space for downstream tasks, as seen in models such as Qwen3-VL~\cite{bai2025qwen3}, InternVL3~\cite{Zhu2025InternVL3EA}, and Video-LLaMA3~\cite{Zhang2023VideoLLaMAAI}. 
Although these techniques enhance offline processing, they depend on full video access and thus struggle in real-world applications where frames arrive sequentially and future observations remain unknown. This has motivated the development of online-oriented models. For example, VideoLLM-Online~\cite{Chen2024VideoLLMonlineOV} was designed to distinguish key frames and redundant frames, Dispider~\cite{Qian2025DispiderEV} utilizes three asynchronous models to allow low response latency, and FlashVStream~\cite{Wu2024VideoLLMMoDEV} and TimeChatOnline~\cite{Yao2025TimeChatOnline8V} aim to handle continuous video input under real-time constraints. While these models perform well on third-person videos and egocentric videos from common daily scenarios, their ability to plan procedures in embodied scenarios remains largely unexamined. In this work, we systematically assess how the current state-of-the-art MLLMs perform in interaction and planning targets, revealing their limitations and offering in-depth analysis to facilitate future research in this direction.

\section{StreamEQA Benchmark}

In this section, we provide a comprehensive introduction to the StreamEQA benchmark. We begin by discussing the design principles and the taxonomy of question-answering tasks, followed by an explanation of the data curation pipeline, and conclude with dataset statistics.

\subsection{Design Principles}
\label{subsec:design_principles}

\begin{figure*}[t]
  \centering
  \includegraphics[
    width=0.8\textwidth,  
    trim=5.0cm 3.5cm 5.0cm 3.5cm,  
    clip
  ]{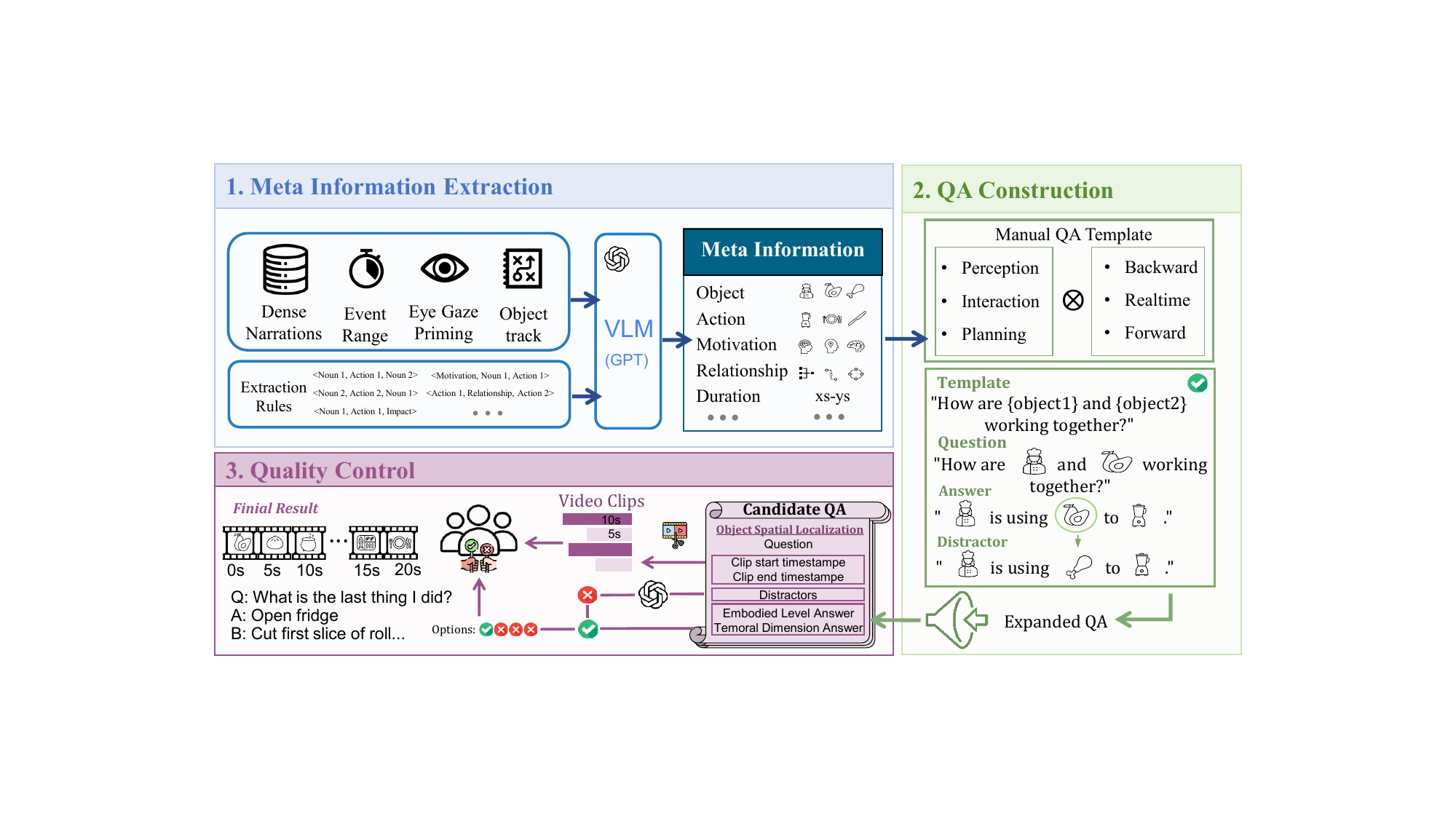}
  \caption{Data construction pipeline of StreamEQA.}
  \label{fig:fig2}
\end{figure*}

The construction of StreamEQA is guided by principles designed to ensure that the benchmark accurately reflects the perceptual and cognitive challenges faced by embodied agents in realistic, continuously evolving real-world environments. These principles emphasize comprehensive embodied task coverage, temporal reasoning across long visual horizons, and context-grounded decision-making. Specifically, StreamEQA is built around the following key design principles.

\noindent
\textbf{Comprehensive Embodied Task Coverage.}
Realistic embodied intelligence depends on the coordinated functioning of several cognitive processes. An effective agent must form stable perceptual representations of surroundings, track the progression of actions and interactions, and reason about goals along with the constraints that govern their achievement. StreamEQA incorporates tasks across this entire spectrum, enabling systematic evaluation of whether a model can recognize object semantics, follow procedural structure, interpret causal dependencies, and construct or refine action plans within an extended activity sequence. This design provides coverage of the full perception–interaction–reflection loop that underlies real-world embodied behavior.

\noindent
\textbf{Streaming-Oriented Temporal Continuity.}
In continuous egocentric video, long-range temporal dependencies are key to understanding dynamic activities. To represent these dependencies explicitly, each task in StreamEQA is formulated under three temporal perspectives: (1) \emph{Backward} perspective evaluates the reconstruction of earlier states, actions, and preconditions; (2) \emph{Realtime} perspective measures understanding of the present visual moment, including object states, spatial organization, and procedural context; (3) \emph{Forward} perspective examines anticipatory reasoning, which involves predicting future events, outcomes, or procedural requirements. These perspectives jointly enforce reasoning that remains coherent across the entire temporal stream and is sensitive to the dynamics of unfolding activities.

\noindent
\textbf{Situated and Causally Grounded Reasoning.}
All questions in StreamEQA are anchored to specific temporal segments of the video and require answers supported directly by the visual and functional structure of the ongoing activity. The formulation prevents reliance on global priors or generic patterns by requiring models to identify the purpose of actions, interpret relationships among manipulated objects, and infer consequences grounded in observable evidence. This principle encourages fine-grained causal interpretation and ensures that model performance reflects genuine comprehension of temporally evolving tasks.

Together, these principles establish StreamEQA as a benchmark for evaluating continuous and context-sensitive embodied cognition. Models assessed under this framework must maintain stable perceptual grounding over time, reason coherently about procedural and causal structure, and generate predictions or plans that align with the constraints and objectives characterizing real-world embodied activity.

\subsection{Task Taxonomy}
\label{subsec:taxonomy}

\begin{figure*}[t]
  \centering
  \includegraphics[
    width=1.0\textwidth,  
    trim=0.6cm 2.5cm 1.0cm 1.0cm,  
    clip
  ]{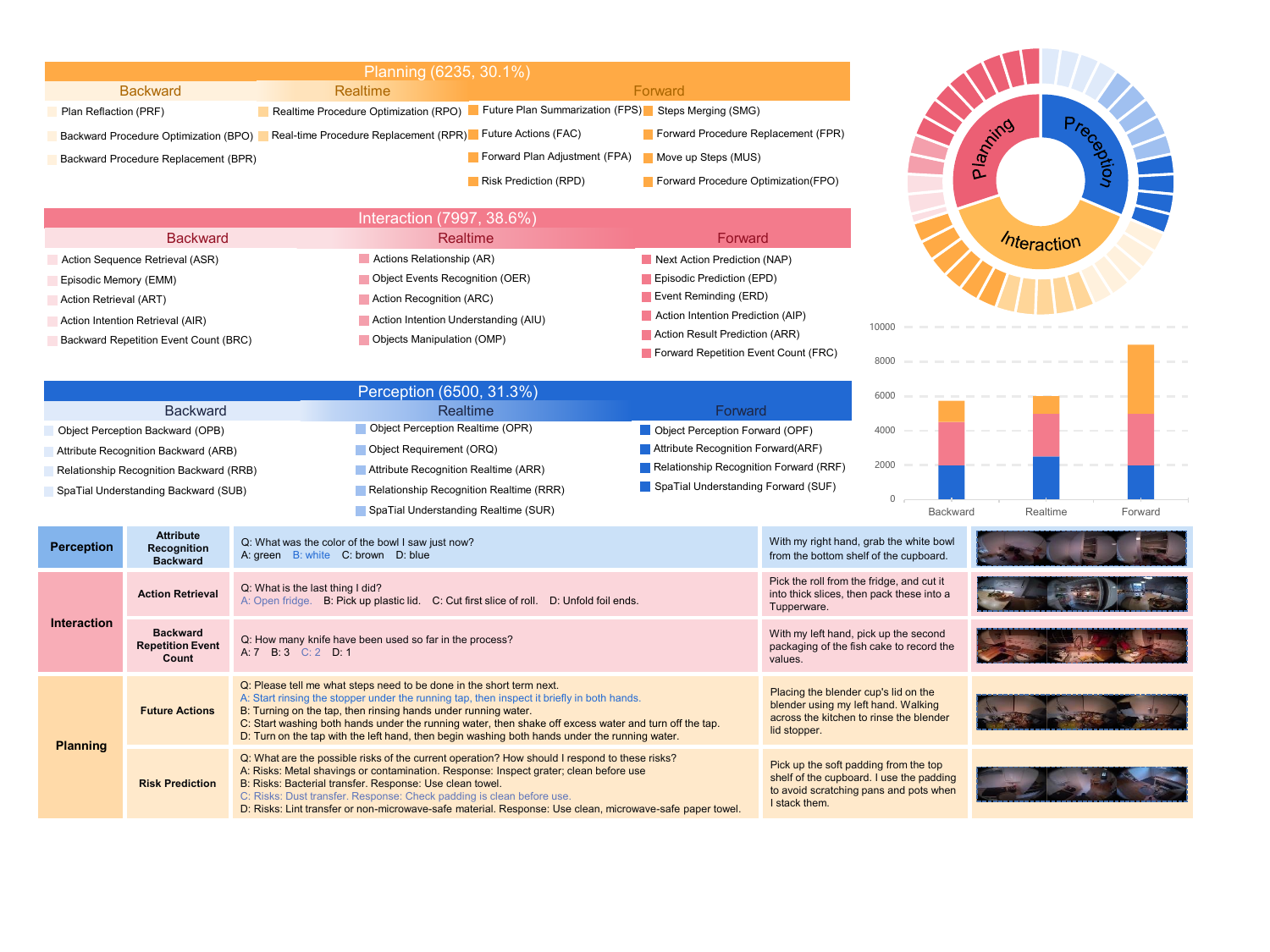}
  \caption{Overview of the StreamEQA task taxonomy and statistics. \textbf{Top-left}: The overall task taxonomy of the three main embodied levels. \textbf{Top-right}: The distributions of questions across embodied levels and temporal dimension. \textbf{Bottom}: A selection of representative QA examples for each major capability.}
  \label{fig:fig3}
\end{figure*}

Based on the above principles, StreamEQA organizes tasks along two orthogonal dimensions.  
The \textbf{Embodied Dimension} reflects the cognitive stage involved: perception, interaction, and planning. The \textbf{Streaming Dimension} reflects the temporal perspective from which the task must be solved.
Crossing these dimensions yields a structured taxonomy of \textbf{42 fine-grained subtasks} as shown in Fig.~\ref{fig:fig3}, covering the full spectrum of streaming embodied video reasoning. Next, we elaborate on the specific taxonomies under these two dimensions.

\noindent
\textbf{Embodied Dimension.}  
This dimension characterizes the cognitive level of the agent, ranging from low-level scene perception to mid-level interaction understanding and high-level task planning:
\begin{itemize}
    \item \emph{Perception} focuses on recognize entities, attributes, and spatial relations in a continuous egocentric view. For example, the model need to recall the color of the bowl observed earlier (\emph{Attribute Recognition Backward}, ARB) or determine the relative position of a spoon in the next moment (\emph{Spatial Understanding Forward}, SUF).
    
    \item \emph{Interaction} tracks actions and object manipulations over time. Typical tasks include recovering the last completed step (\emph{Action Retrieval Backward}, ARB) or counting how many times a repeated action (\emph{Backward Repetition Event Count}, BRC).

    \item \emph{Planning} evaluates the ability to synthesize or revise plans. Examples include summarizing upcoming steps required to complete the task (\emph{Future Actions}, FAC) or identifying operational risks and proposing prevention strategies (\emph{Risk Prediction}, RPD).
\end{itemize}

\noindent
\textbf{Streaming Dimension.}  
To capture the temporal continuity inherent to streaming video, each embodied level is instantiated under three temporal scopes. This temporal structuring ensures that StreamEQA capable of assessing whether models can sustain temporal coherence and causal consistency over continuous first-person video.
\begin{itemize}
    \item \emph{Backward Reasoning} requires reconstructing earlier states, actions, or causal prerequisites. Tasks under this scope evaluate whether the model can retrieve information no longer visible in the current frame. For example, the model needs to recall objects that appeared in a certain location (\emph{Object Perception Backward}, OPB).

    \item \emph{Realtime Understanding} focuses on moment-to-moment situational awareness within the ongoing egocentric stream. These tasks require maintaining continuous perceptual grounding and recognizing the current procedural context. A central example is the \emph{Spatial Understanding Realtime} (SUR) task, where the model must determine the present relative configuration of objects.

    \item \emph{Forward Prediction} examines the model’s capacity to anticipate future events, outcomes, or required operations. This temporal scope probes prospective inference, a key component of embodied planning. A representative instance is the \emph{Risk Prediction} (RPD) task, where the model needs to predict potential risks.
\end{itemize}
This taxonomy provides a principled organization of streaming embodied reasoning. By combining embodied stages (Perception, Interaction, Planning) with temporal scopes (Backward, Realtime, Forward), StreamEQA systematically evaluates whether multimodal models can integrate memory, situational awareness, and foresight in continuous egocentric videos.

\subsection{Data Construction}
\label{sec:data_construction}

We construct the StreamEQA dataset through a three-stage pipeline to generate temporally grounded question–answer pairs from long egocentric videos. 
As illustrated in Fig.~\ref{fig:fig2}, the process consists of \emph{Meta Information Extraction}, \emph{QA Construction}, and \emph{Quality Control}. 
This design ensures that each question is generated from semantically grounded video narrations while maintaining temporal precision.

\noindent
\textbf{Meta Information Extraction.}
The first stage involves extracting detailed meta information from meta annotations, sourced from the HD-EPIC dataset, including dense narrations with precise timestamps, event time range, eye gaze priming, and object spatial track.
For each video, a language foundation model (\emph{i.e.}, GPT-5\cite{openai2025gpt5}) processes the diverse meta annotations together with the activity annotations to generate structured meta elements, including \emph{objects, actions, object episoids, object relationships, action motivations,} and \emph{event durations}.  
These meta-informations are aligned with timestamps of narrations and associated events, forming representations of what appears, what occurs, why it happens, and how it works.  
This structured abstraction serves as the foundation for subsequent question synthesis, ensuring that all generated QA items remain grounded in the actual visual and linguistic content. 

\noindent
\textbf{QA Construction.}
Building upon the extracted meta-informations, the second stage generates question–answer pairs under a manually defined schema. We design a set of task-specific question templates based on three embodiment levels: \textit{Perception, Interaction,} and \textit{Planning}. Each embodied level is further divided into three temporal dimensions: \textit{Backward}, \textit{Realtime}, and \textit{Forward}. These produce nine major QA categories (\(3\times3\)), with each category encompassing multiple specific sub-tasks.  

For each template, the placeholders are automatically filled with the relevant types of meta-informations. For example, a backward-perception task template ``\emph{How are \{object1\} and \{object2\} working together just now?}'' is expanded into a grounded question by inserting selected objects (e.g., \emph{cup's content} and \emph{blender}) and actions from the corresponding meta-informations. Meanwhile, the selected meta informations are also composed with the answer template (``\emph{The person is using the \{object1\} to \{action\} the \{object1\}}'') and generate the grounded answer: ``\emph{The person is using the blender to mix the cup’s contents}''. Additionally, while retaining the subject and object, we randomly select actions from the meta-information to replace the correct action (e.g., replace \emph{mix} with \emph{wash}), generating extra 3 distractors with the same context but different details, forming a four-choice QA pair. Finally, the query time is set after the timestamp of the event to simulate the scenario of backward tracing. The construction pipeline guarantees that each QA pair is supported by grounded meta annotations.

\noindent
\textbf{Quality Control.}
To enhance the challenges of the QA pairs, we leverage GPT-5\cite{openai2025gpt5} to refine the distractor options. Specifically, GPT-5 is tasked with refining only the incorrect options, while ensuring the correct option remains unchanged. Besides, all options must be related to the question and visual context, ensuring the refined incorrect options are logically plausible, but will ultimately be excluded due to contradiction with video content. Finally, we assessed the clarity of the questions and the ambiguity of the options to prevent misinterpretations, and human annotators verified the accuracy of annotations. The final dataset consists of validated QA triplets ( questions, options, and correct answers) and precise temporal boundaries.

\subsection{Data Statistics}
Our StreamEQA benchmark encompasses real-world egocentric videos and corresponding question-answer annotations, organized into three embodied dimensions and three temporal dimensions. It comprises 156 videos and approximately 21K QA pairs, spanning 42 sub-task types grouped into three embodied levels. Table~\ref{tab:table1} summarizes key statistics of the five datasets, including the number of sub-tasks, QA pairs, and average seconds of video durations (Dur.(s)). More detailed statistical breakdowns and representative examples can be found in Figure~\ref{fig:fig3}.

\begin{table}[t]  
  \centering
  \small
  \adjustbox{max width=\columnwidth}{
  \begin{tabular}{l|ccc}  
    \toprule
    Tasks & Sub-Task & QA Pairs & Duration(s) \\
    \midrule
    Perception & 13 & 6,500 & 2.43 \\
    Interaction & 16 & 7,996 & 112.67 \\
    Planning & 13 & 6,235 & 91.67 \\
    \midrule
    Backward & 12 & 5,736 & 69.75 \\
    Realtime & 12 & 5,996 & 34.73 \\
    Forward & 18 & 8,499 & 88.25 \\
    \midrule
    StreamEQA & 42 & 20,731 & 67.55 \\
    \bottomrule
  \end{tabular}
  }
  \caption{Key statistics of StreamEQA benchmark.}
  \label{tab:table1}
\end{table}

\section{Experiments}
This section presents a set of comprehensive experiments and in-depth analysis of our proposed StreamEQA. We first outline the experimental settings in Section \ref{sec:4_1}. The main results of state-of-the-art MLLMs are then presented in Section \ref{sec:4_2}. In Section \ref{sec:4_3}, we conduct a comparative analysis with previous benchmarks through quantitative experiments to further validate the challenge posed by StreamEQA. Finally, Section \ref{sec:4_4} explores the gap between the offline-online settings. 

\subsection{Models and Evaluation Settings}
\label{sec:4_1}

\begin{table*}[!t]
\centering
\resizebox{\linewidth}{!}{
\begin{tabular}{l|cc|c|cccc|cccc|cccc|c}
\toprule
\multirow{2}{*}{\textbf{Models}} & \multirow{2}{*}{\textbf{OS}} & \multirow{2}{*}{\textbf{Ego}} & \multirow{2}{*}{\textbf{Fre.}} & \multicolumn{4}{c|}{\textbf{Perception}} & \multicolumn{4}{c|}{\textbf{Interaction}} & \multicolumn{4}{c|}{\textbf{Planning}} & \textbf{Avg.} \\
& & & & Bac. & Rea. & For. & Avg. & Bac. & Rea. & For. & Avg. & Bac. & Rea. & For. & Avg. & \\
\midrule
\multicolumn{16}{c}{\textit{Offline Video Understanding MLLMs}} \\
\midrule
Gemini & \XSolidBrush & \XSolidBrush & 1fps & \textbf{71.4} & \textbf{68.2} & 71.3 & \textbf{70.1} & 58.7 & 59.2 & 52.1 & 56.4 & 52.9 & 54.4 & \textbf{51.8} & 52.5 & 60.0\\
GPT5 & \XSolidBrush & \XSolidBrush & 1fps & 70.1 & 65.8 & \textbf{73.2} & 69.3 & 60.8 & 62.2 & \textbf{56.9} & 59.8 & 67.2 & 57.4 & 51.2 & 55.3 & \textbf{61.3}\\
InternVL3 & \Checkmark & \XSolidBrush & 64 & 63.5 & 62.1 & 65.2 & 63.5 & 42.1 & 42.3 & 37.3 & 40.4 & 50.8 & 47.7 & 44.9 & 46.51 & 49.5 \\
LongVA & \Checkmark & \XSolidBrush & 64 & 58.7 & 55.8 & 57.3 & 57.2 & 28.5 & 27.4 & 24.0 & 26.5 & 36.9 & 39.4 & 31.5 & 33.8 & 38.4 \\
MiniCPM-V & \Checkmark & \XSolidBrush & 64 & 64.8 & 63.0 & 64.3 & 63.9 & 40.4 & 35.1 & 47.2 & 41.0 & 47.0 & 39.5 & 36.5 & 39.2 & 47.8 \\
Qwen3VL & \Checkmark & \XSolidBrush & 64 & 65.1 & 58.9 & 66.0 & 63.0 & 53.5 & 49.2 & 47.3 & 49.9 & 50.0 & 43.5 & 39.3 & 42.0 & 51.7 \\
VideoLLaMA3 & \Checkmark & \XSolidBrush & 64 & 59.2 & 58.0 & 57.1 & 58.1 & 36.3 & 34.7 & 29.4 & 33.2 & 38.3 & 37.6 & 36.1 & 36.8 & 42.1 \\
EgoGPT & \Checkmark & \Checkmark & 64 & 67.4 & 64.3 & 67.8 & 66.3 & \textbf{62.7} & \textbf{64.9} & 55.1 & \textbf{60.5} & \textbf{67.3} & \textbf{58.3} & 51.6 & 55.8 & 60.9 \\
EgoVLPv2 & \Checkmark & \Checkmark & 64 & 23.6 & 25.1 & 23.8 & 24.2 & 22.9 & 25.5 & 19.3 & 22.4 & 23.8 & 25.1 & 25.1 & 24.8 & 23.7 \\
\midrule
\multicolumn{16}{c}{\textit{Online Video Understanding MLLMs}} \\
\midrule
VideollmOnline & \Checkmark & \XSolidBrush & 64 & 36.2 & 38.4 & 33.2 & 36.1 & 32.8 & 31.4 & 26.7 & 30.2 & 36.0 & 35.3 & 38.1 & 37.2 & 34.1 \\
FlashVStream & \Checkmark & \XSolidBrush & 64 & 62.0 & 57.3 & 60.1 & 59.6 & 45.8 & 40.6 & 36.4 & 40.7 & 49.4 & 47.3 & 43.4 & 45.2 & 48.0 \\
Dispider & \Checkmark & \XSolidBrush & 64 & 57.5 & 56.2 & 53.8 & 55.9 & 29.2 & 35.3 & 24.8 & 29.5 & 53.5 & 46.3 & 43.5 & 45.9 & 42.8 \\
TimeChatOnline & \Checkmark & \XSolidBrush & 64 & 46.6 & 48.7 & 48.4 & 47.9 & 34.0 & 28.0 & 25.8 & 29.1 & 42.1 & 40.7 & 32.4 & 35.6 & 36.9 \\
\bottomrule
\end{tabular}
}
\caption{\textbf{Detailed evaluation results on StreamEQA}. All scores are reported in percentages. The best results are marked in bold, and the second-best are underlined. OS: Open-sourced Models. Ego: Egocentic Models. Fra: Frames. Bac: Backward. Rea: Realtime. For: Forward. Avg: Average.}
\label{tab:table2}
\end{table*}

\begin{figure*}[t]
  \centering
  \includegraphics[
    width=1.0\textwidth,  
    trim=0.3cm 5.5cm 0.8cm 5.3cm,  
    clip
  ]{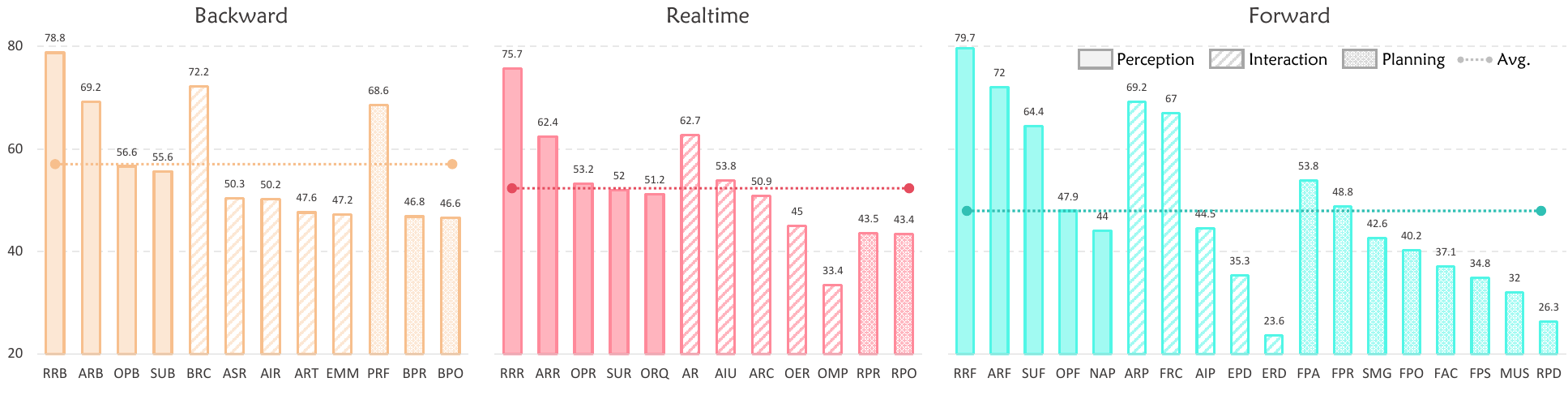}
  \caption{The performence of Qwen3VL on 42 Online-Embodied tasks which are categorized by the temporal dimension, while sorting tasks by accuracy within the same embodied level.}
  \label{fig:fig4}
\end{figure*}

\paragraph{Evaluated Models.} 
We select a diverse set of MLLMs spanning three categories to cover major technical paradigms: (1) Proprietary Models, including GPT-5 \cite{openai2025gpt5}, Gemini-2.5-Pro \cite{comanici2025gemini}, (2) Offline Video Understanding MLLMs, including Qwen3VL-8B \cite{bai2025qwen3}, LongVA-7B \cite{Zhang2024LongCT}, MiniCPM-V-4.5-8B \cite{Yu2025MiniCPMV4C}, InternVL3-8B \cite{Zhu2025InternVL3EA} and VideoLLaMA3-7B \cite{Zhang2023VideoLLaMAAI}, (2) Online Video Understanding MLLMs, including FlashVStream-7B \cite{Wu2024VideoLLMMoDEV}, VideollmOnline-8B \cite{Chen2024VideoLLMonlineOV}, Dispider-7B \cite{Qian2025DispiderEV} and TimeChatOnline-8B \cite{Yao2025TimeChatOnline8V}, (3) Egocentric MLLMs, including EgoGPT-7B \cite{yang2025egolife}, EgoVLPv2 \cite{Pramanick2023EgoVLPv2EV}. 

\paragraph{Evaluation and Implementation Details.} 
Following prior works\cite{Li2025OVOBenchHF,Lin2024StreamingBenchAT,li2025egocross}, we use the standard accuracy as our evaluation metric, which is calculated as the percentage of correctly answered questions. To ensure a fair comparison of model performance, we adhere to the principle of consistency by maintaining the same number of frames or frames per second (fps) across all models. Considering the input video length limitations for offline Video-LLMs, we adopt specialized video input methods tailored to such models. Specifically, we segment the video into clips based on the time range of the event. For example, for an event starting at timestamp $t_e$ and a question $Q_i$ being posed at timestamp $t_q$, we extract the video clip $Video[t_e: t_q]$ as visual input. This approach simulates a streaming question-answering scenario in online video understanding while reducing information loss of relative events in related video clips. 
We also conduct a pilot study of Qwen3-VL-8B \cite{bai2025qwen3} and EgoGPT \cite{yang2025egolife}. In this setting, we re-evaluate the two models on the whole event clips to simulate the conventional offline QA, and then plot the change of accuracy on three embodied-level tasks.
All experiments are conducted on NVIDIA A6000 GPUs.

\subsection{Main Results}
\label{sec:4_2}
Table~\ref{tab:table2} reports the performance of 13 MLLMs on StreamEQA, covering Perception, Interaction, and Planning, as well as their overall performance. Our evaluation brings several important findings as follows:

\noindent
\textbf{Overall Difficulty.} All MLLMs evaluated on StreamEQA demonstrate significant challenges across the entire benchmark. The best-performing model, GPT-5, achieves an overall accuracy of 61.3\%, while the lowest-performing model, EgoVLPv2, scores only 23.7\%. Most mainstream models fluctuate around 40\%, offering no significant advantage over the 25\% random guessing accuracy. These results highlight the intrinsic difficulty of the benchmark, indicating that even the state-of-the-art models face considerable challenges in the streaming embodied video understanding tasks posed by StreamEQA.

\noindent
\textbf{Embodied-Level Comparison.}
Across different embodied levels, we observe varying degrees of difficulty: perception is relatively easier, followed by interaction, with planning being the most challenging. Taking Qwen3VL as an example, its performance in perception is 63.0\%, in interaction it drops to 49.9\%, and in planning it further decreases to 42.0\%. This performance decline is consistent across models, highlighting a notable reduction in accuracy from perception to planning. This outcome aligns with expectations, as the predominant training paradigm for mainstream MLLMs focuses on object and scene perception tasks. In contrast, we were surprised by the superior performance of online Video-LLMs in planning tasks compared to interaction tasks. For instance, FlashVStream achieves 40.7\% in interaction, but performs better in planning with 45.2\%. This suggests that the training process for online models may better align with the dynamic, task-oriented planning requirements often seen in real-world scenarios.

\noindent
\textbf{Model-wise Performance.} Despite the two proprietary MLLMs, EgoGpt and Qwen3VL achieve the highest overall average accuracy among offline Video-MLLMs, with EgoGpt (60.9\%) significantly exceeding Qwen3VL (51.7\%). We assume that, as the embodied-specific model, EgoGPT was explicitly designed and trained on embodied video data, which makes it outperform especially in interaction (60.5\%) and planning (55.8\%), and even performs close to GPT-5. In addition, online Video-LLMs have relatively lower performance compared with offline Video-LLMs, which indicates that the evaluated online MLLMs are not effective in embodied scenarios though they are well designed for online video understanding.

\subsection{Temporal-level Variance} 
\label{sec:4_3}
As illustrated in Figure~\ref{fig:fig4}, we take Qwen3VL as an example to further analyze the results across different temporal dimensions, including backward, realtime, and forward. Since MLLMs can acquire sufficient past and current information from the video stream, the model naturally achieves higher accuracy in backward and realtime tasks compared to forward tasks. 
Additionally, we observe that backward scores tend to be more stable, while forward tasks are more sensitive to variations. 
These findings further suggest that MLLMs face significant challenges in anticipating future events and planning accordingly.

\subsection{{More In-Depth Analysis}}   
We conduct further studies to explore and quantify the distinct challenges introduced by the streaming and embodied nature of the StreamEQA benchmark.

\label{sec:4_4}
\begin{figure}[t]
  \centering
  \includegraphics[
    width=1.0\columnwidth,
    trim=6.1cm 4.0cm 6.7cm 4.0cm,
    clip
    ]{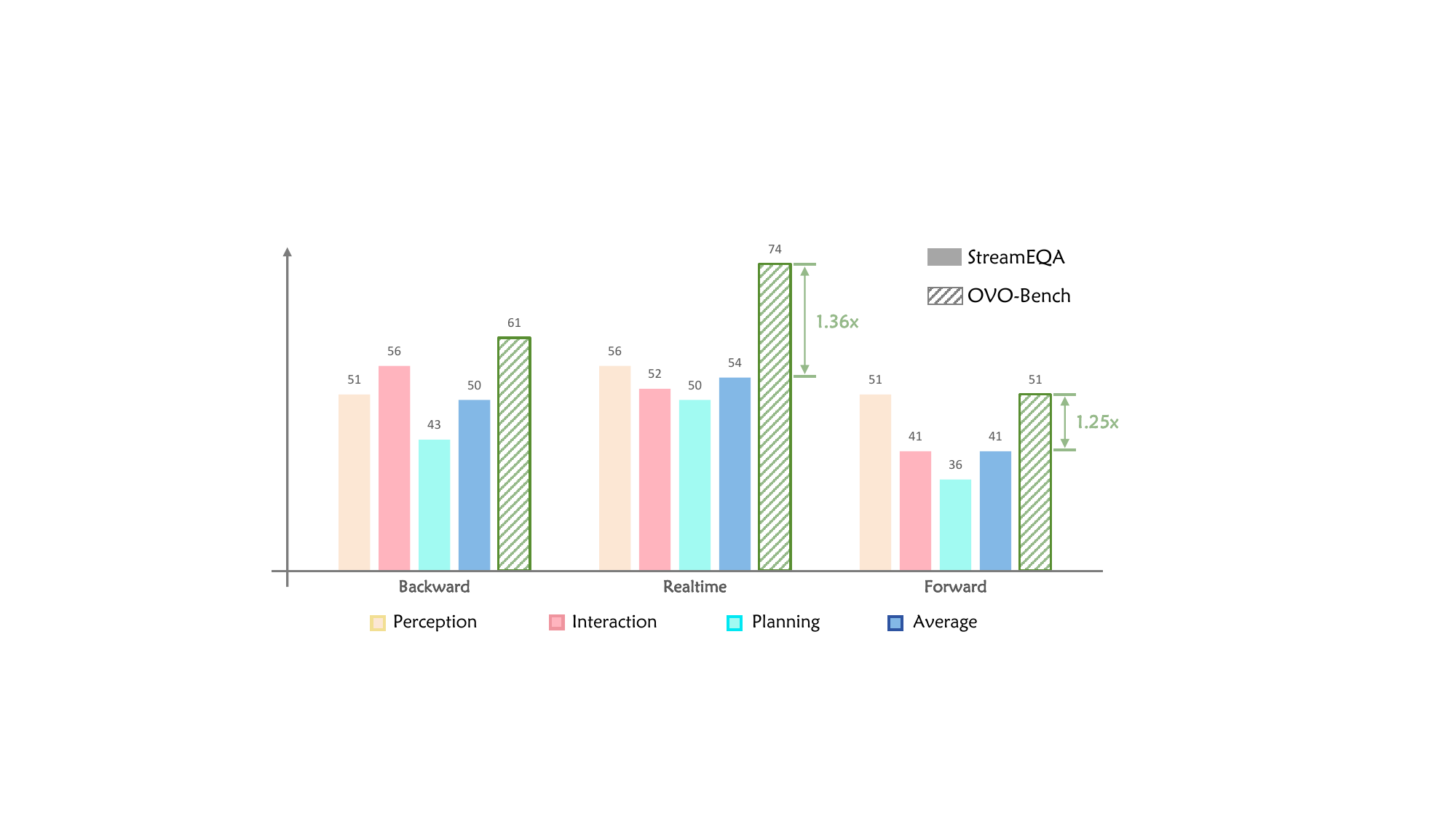}  
  \caption{Online and Online-Embodied performance comparison. The results highlight the gap caused by embodied scenarios.}
  \label{fig:fig5}
\end{figure}

\noindent
\textbf{Embodied Dimension Challenges.} To further examine embodied tasks and their challenges, we compare the performance of Qwen3VL on our benchmark and OVO-bench, a typical streaming video dataset for general scenario.
Results in Figure~\ref{fig:fig5} reveal a consistent and significant performance drop across all comparable QA types. For instance, performance drops from an impressive 74\% on realtime tasks in OVO-bench to just 54.26\% on the realtime embodied tasks in StreamEQA. Similarly, the accuracy of forward prediction tasks falls from 51.51\% to 41.18\%, quantifying the substantial penalty incurred by the embodied tasks.

\begin{table}[t]  
  \centering
  \small
  \adjustbox{max width=0.9\columnwidth}{
  \begin{tabular}{l|c|ccc|c}  
    \toprule
    Method & Settings & Bac. & Rea. & For. & Avg. \\
    \midrule
    \multirow{2}{*}{EgoGPT} & offline & 65.7 & 65.83 & 57.22 & 62.03 \\
    & online & 65.29 & 63.54 & 56.4 & 60.93 \\
    \bottomrule
  \end{tabular}
  }
  \caption{Comparison between online and offline settings of StreamEQA.}
  \label{tab:table3}
  \vspace{-0.15in}
\end{table}

\noindent
\textbf{Streaming Dimension Challenges.} As shown in Table~\ref{tab:table3}, we further explore the challenges of streaming video inputs setting. Instead of extracting video clips before query timestamps, models in offline settings can get complete clips of events to simulate the conventional evaluation. We evaluate the performance of EgoGPT on StreamEQA in embodied and online-embodied settings separately. The results indicate that EgoGPT, as an embodied video-MLLM trained in offline settings, performs better at embodied settings, while having lower accuracy when inferring on online-embodied settings (nearly 1.10\% reduction). The results indicate that streaming video inputs contribute to the challenges in embodied scenarios.

\section{Conclusion}
In this work, we present StreamEQA, a new benchmark for evaluating the ability of MLLMs in online-embodied video question answering. StreamEQA comprises approximately 21K QA pairs systematically generated based on meta annotations and curated from three embodied levels: perception, interaction, and planning. It spans a wide range of QA types, enabling fine-grained evaluation. Beyond constructing the benchmark, we extensively evaluate current state-of-the-art MLLMs and find that, despite strong performance on perceiving objects in videos, they struggle to conduct interaction and planning in embodied scenarios. Additionally, we further separately explore the challenges of streaming video input and embodied tasks. We believe that StreamEQA, together with our experiments and analysis, offers a valuable foundation for future research on robust and practically applicable embodied streaming video understanding.

{
    \small
    \bibliographystyle{ieeenat_fullname}
    \bibliography{main}
}

\clearpage
\maketitlesupplementary

\begin{table*}[h]
\centering
\begin{tabular}{l c c c c c c}
\toprule
\textbf{Dataset} & \textbf{Video Length} & \textbf{\# Test} & \textbf{\# Categories} & \textbf{Streaming} & \textbf{Embodied} \\
\midrule
Video-MME~\cite{Fu2024VideoMMETF} & 180s & 2.7K & 12 & \ding{55} & \ding{55} \\
EgoThink~\cite{cheng2024egothink} & - & 750  & 12   & \ding{55} & \checkmark \\
EgoSchema~\cite{Mangalam2023EgoSchemaAD} & 3 min       & 500 & -   & \ding{55} & \checkmark \\
StreamingBench~\cite{Lin2024StreamingBenchAT} & (3s, 24 min) & 4.5K & 18  & \checkmark & \ding{55} \\
OVO-Bench~\cite{Li2025OVOBenchHF} & 7 min & 2.8K & 12   & \checkmark & \ding{55} \\
\midrule
\textbf{StreamEQA} & 67.5s & 21K & 42  & \checkmark & \checkmark \\
\bottomrule
\end{tabular}
\caption{Overview of each dataset's characteristics, including average video length, number of test examples, number of categories, and question types (Streaming or Embodied).}
\label{tab:table6}
\end{table*}

\begin{table*}[h]
\centering
\resizebox{\linewidth}{!}{
\begin{tabular}{l|c|cccc|ccccc|cccc|c}
\toprule
\multirow{2}{*}{\textbf{Models}} & \multirow{2}{*}{\textbf{Frames}} & \multicolumn{4}{c|}{\textbf{Backward}} & \multicolumn{5}{c|}{\textbf{Realtime}} & \multicolumn{4}{c|}{\textbf{Forward}}  & \multirow{2}{*}{\textbf{Avg.}} \\
& & OPB & ARB & RRB & SUB & OPR & ORQ & ARR & RRR & SUR & OPF & ARF & RRF & SUF \\
\midrule
\multicolumn{16}{c}{\textit{Offline Video Understanding MLLMs}} \\
\midrule
Gemini & 1fps & 64.2 & \textbf{76.3} & \textbf{82.8} & 62.0 & 67.1 & 58.7 & \textbf{76.1} & \textbf{82.4} & \textbf{56.4}& 58.4 & 79.6 & 82.6 & 64.7 & \textbf{70.1} \\
GPT5 & 1fps & 60.4 & 75.8 & 83.2 & 53.6 & 61.4 & 58.6 & 73.4 & 83.5 & 52.2& 60.3 & \textbf{80.6} & \textbf{84.5} & \textbf{65.7} & 69.3 \\
InternVL3 & 64 & 60.8 & 71.8 & 70.2 & 51.0 & 63.8 & 58.8 & 71.2 & 70.7 & 45.8& \textbf{62.1} & 73.8 & 73.3 & 51.6 & 63.5 \\
LongVA & 64 & 49.0 & 58.4 & 64.8 & \textbf{62.4} & 48.8 & 46.6 & 59.2 & 67.5 & 57.0& 41.1 & 63.8 & 68.5 & 56.0 & 57.2 \\
MiniCPM-V & 64 & 63.0 & 68.6 & 70.6 & 56.8 & 62.8 & \textbf{62.4} & 67.0 & 71.9 & 50.8& 61.7 & 62.8 & 77.9 & 54.6 & 63.9 \\
Qwen3VL & 64 & 56.6 & 69.2 & 78.8 & 55.6 & 53.2 & 51.2 & 62.4 & 75.7 & 52.0& 47.9 & 72.0 & 79.7 & 64.5 & 63.0 \\
VideoLLaMA3 & 64 & 58.4 & 65.8 & 63.8 & 48.8 & 58.2 & 53.8 & 66.8 & 63.5 & 47.6& 53.3 & 56.2 & 65.7 & 53.2 & 58.1 \\
EgoGPT & 64 & \textbf{69.0} & 69.2 & 69.6 & 61.6 & \textbf{67.2} & 61.2 & 66.0 & 70.5 & \textbf{56.4}& 58.7 & 75.4 & 72.5 & 64.5 & 66.3 \\
EgoVLPv2 & 64 & 22.4 & 22.4 & 27.2 & 22.4 & 25.2 & 26.6 & 25.8 & 25.5 & 22.4& 23.9 & 21.8 & 23.5 & 26.1 & 24.2 \\
\midrule
\multicolumn{16}{c}{\textit{Online Video Understanding MLLMs}} \\
\midrule
Dispider & 64 & 56.4 & 66.6 & 59.6 & 47.2 & 59.2 & 55.6 & 64.6 & 61.3 & 40.2& 56.9 & 58.0 & 59.8 & 40.6 & 55.9 \\
FlashVStream & 64 & 55.8 & 65.2 & 69.2 & 57.8 & 51.2 & 50.0 & 62.6 & 68.9 & 53.6& 49.7 & 62.8 & 71.08 & 56.6 & 59.6 \\
TimeChatOnline & 64 & 38.8 & 46.4 & 56.8 & 44.4 & 40.2 & 45.4 & 45.8 & 63.1 & 48.8& 38.9 & 53.0 & 62.9 & 38.8 & 47.9 \\
VideollmOnline & 64 & 28.2 & 43.4 & 32.6 & 40.4 & 25.8 & 28.4 & 40.8 & 39.5 & 57.4& 28.7 & 20.6 & 39.0 & 44.8 & 36.1 \\
\bottomrule
\end{tabular}
}
\caption{Evaluation on Sub-tasks about Perception.}
\label{tab:table8_1}
\end{table*}

\begin{table*}[h]
\centering
\resizebox{\linewidth}{!}{
\begin{tabular}{l|c|ccccc|ccccc|cccccc|c}
\toprule
\multirow{2}{*}{\textbf{Models}} & \multirow{2}{*}{\textbf{Frames}} & \multicolumn{5}{c|}{\textbf{Backward}} & \multicolumn{5}{c|}{\textbf{Realtime}} & \multicolumn{6}{c|}{\textbf{Forward}} & \multirow{2}{*}{\textbf{Avg.}} \\
& & ASR & EMM & ART & AIR & BRC & AR & OER & ARC & AIU & OMP & NAP & EPD & ERD & AIP & ARP & FRC \\
\midrule
\multicolumn{19}{c}{\textit{Offline Video Understanding MLLMs}} \\
\midrule
Gemini & 1fps & 55.2 & 73.7 & 51.9 & 64.6 & 44.8& 63.1 & 49.7 & 60.7 & 61.6 & \textbf{62.2} & 37.2 & 60.2 & 37.7 & 57.0 & 73.0 & 45.5 & 56.4\\
GPT5 & 1fps & \textbf{62.1} & 72.4 & 50.0 & 71.8 & 47.8& \textbf{71.3} & 49.6 & 61.2 & \textbf{76.6} & 52.2& 38.0 & \textbf{66.7} & 39.6 & \textbf{67.9} & 84.9 & 44.4 & 59.8\\
InternVL3 & 64 & 45.5 & 44.9 & 39.8 & 48.0 & 32.3& 60.5 & 36.6 & 36.9 & 44.2 & 32.9& 37.4 & 36.5 & 16.8 & 41.1 & 65.8 & 26.2 & 40.3\\
LongVA & 64 & 43.1 & 14.9 & 23.0 & 26.1 & 35.4& 58.5 & 8.6 & 10.0 & 27.4 & 32.9& 37.6 & 15.7 & 12.3 & 22.0 & 41.3 & 15.2 & 26.5\\
MiniCPM-V & 64 & 43.1 & 41.5 & 43.8 & 40.2 & 55.8& 58.6 & 33.6 & 30.7 & 39.4 & 52.0& 34.4 & 33.3 & 9.1 & 31.9 & 62.6 & 41.6 & 39.9\\
Qwen3VL & 64 & 50.3 & 47.3 & 47.6 & 50.2 & \textbf{72.2}& 62.7 & 45.0 & 50.9 & 53.8 & 33.5& \textbf{44.0} & 35.3 & 23.6 & 44.6 & 69.2 & \textbf{67.0} & 49.9\\
VideoLLaMA3 & 64 & 36.9 & 29.2 & 30.0 & 32.2 & 53.2& 60.9 & 19.6 & 27.9 & 35.2 & 30.0& 33.0 & 17.1 & 11.9 & 26.4 & 42.05 & 45.8 & 33.2\\
EgoGPT & 64 & 43.1 & \textbf{78.3} & \textbf{59.0} & \textbf{72.4} & 60.8& 60.1 & \textbf{73.0} & \textbf{84.3} & 68.8 & 37.8& 40.6 & 59.0 & \textbf{49.5} & 61.7 & \textbf{85.7} & 34.2 & \textbf{60.5}\\
EgoVLPv2 & 64 & 10.7 & 26.5 & 26.9 & 26.6 & 24.1& 28.3 & 26.2 & 21.5 & 26.8 & 25.1& 11.4 & 25.3 & 2.4 & 25.8 & 26.8 & 24.2 & 22.4\\
\midrule
\multicolumn{19}{c}{\textit{Online Video Understanding MLLMs}} \\
\midrule
Dispider & 64 & 29.4 & 26.1 & 38.8 & 36.5 & 15.7& 63.4 & 24.5 & 27.6 & 34.8 & 26.4& 28.0 & 20.2 & 6.1 & 30.0 & 57.4 & 10.7 & 29.5\\
FlashVStream & 64 & 49.1 & 36.5 & 33.5 & 44.2 & 65.2& 64.6 & 27.4 & 39.0 & 44.6 & 27.3& 36.5 & 19.6 & 4.2 & 35.5 & 63.1 & 56.4 & 40.7\\
TimeChatOnline & 64 & 48.1 & 18.2 & 23.6 & 23.0 & 56.8& 62.5 & 15.0 & 17.9 & 20.4 & 24.0& 40.4 & 13.1 & 15.0 & 17.9 & 39.7 & 28.4 & 29.5\\
VideollmOnline & 64 & 26.7 & 33.0 & 17.8 & 38.2 & 42.2& 57.3 & 30.2 & 24.5 & 37.8 & 6.9& 34.6 & 22.5 & 0.2 & 23.6 & 50.5 & 36.6 & 30.2\\
\bottomrule
\end{tabular}
}
\caption{Evaluation on Sub-tasks about Interaction.}
\label{tab:table8_2}
\end{table*}

\begin{table*}[h]
\centering
\resizebox{\linewidth}{!}{
\begin{tabular}{l|c|ccc|cc|cccccccc|c}
\toprule
\multirow{2}{*}{\textbf{Models}} & \multirow{2}{*}{\textbf{Frames}} & \multicolumn{3}{c|}{\textbf{Backward}} & \multicolumn{2}{c|}{\textbf{Realtime}} & \multicolumn{8}{c|}{\textbf{Forward}} & \multirow{2}{*}{\textbf{Avg.}} \\
& & PRF & BPO & BPR & RPO & RPR & FPS & FAC & FPA & RPD & SMG & FPR & MUS & FPO \\
\midrule
\multicolumn{16}{c}{\textit{Offline Video Understanding MLLMs}} \\
\midrule
Gemini & 1fps & 65.8 & 53.6 & 51.2 & 54.5 & 54.3 & \textbf{46.7} & 47.9 & 55.3 & \textbf{34.4} & 53.3 & 52.7 & 49.2 & 60.7 & 52.5\\
GPT5 & 1fps & \textbf{81.5} & 64.6 & 64.6 & \textbf{61.0} & 53.8 & 45.4 & \textbf{48.9} & \textbf{62.1} & 30.7 & 59.1 & 46.8 & 50.0 & \textbf{67.7} & 55.3\\
InternVL3 & 64 & 73.0 & 41.6 & 51.8 & 47.2 & 48.2 & 40.0 & 43.3 & 59.6 & 28.9 & 54.7 & 52.4 & 38.8 & 43.4 & 46.5\\
LongVA & 64 & 53.0 & 34.01 & 33.9 & 42.2 & 36.5 & 22.6 & 27.3 & 50.1 & 20.0 & 29.7 & 33.5 & 37.9 & 33.1 & 33.8\\
MiniCPM-V & 64 & 63.5 & 42.0 & 47.5 & 39.8 & 39.7 & 33.3 & 33.8 & 48.7 & 26.3 & 34.9 & 41.6 & 35.6 & 37.6 & 39.2\\
Qwen3VL & 64 & 68.7 & 46.6 & 46.8 & 43.4 & 43.6 & 34.8 & 37.1 & 53.9 & 26.3 & 42.7 & 48.8 & 32.0 & 40.2 & 42.0\\
VideoLLaMA3 & 64 & 55.1 & 28.8 & 41.6 & 34.6 & 40.6 & 33.4 & 35.9 & 44.3 & 27.3 & 39.1 & 41.4 & 32.2 & 35.9 & 36.7\\
EgoGPT & 64 & 62.7 & \textbf{67.8} & \textbf{68.4} & 53.8 & \textbf{62.8} & 42.0 & 40.7 & 59.8 & 31.1 & \textbf{62.5} & \textbf{74.1} & \textbf{50.4} & 53.4 & \textbf{55.7}\\
EgoVLPv2 & 64 & 27.7 & 22.4 & 23.8 & 23.0 & 27.1 & 26.4 & 20.0 & 25.5 & 23.0 & 28.3 & 24.5 & 25.4 & 27.5 & 24.8\\
\midrule
\multicolumn{16}{c}{\textit{Online Video Understanding MLLMs}} \\
\midrule
Dispider & 64 & 59.7 & 44.1 & 60.5 & 37.9 & 54.5 & 31.2 & 34.4 & 46.5 & 30.6 & 53.9 & 65.1 & 48.2 & 35.7 & 45.9\\
FlashVStream & 64 & 69.7 & 39.3 & 51.8 & 46.3 & 48.2 & 37.3 & 40.8 & 59.8 & 22.1 & 47.5 & 52.7 & 43.5 & 45.8 & 45.2\\
TimeChatOnline & 64 & 55.2 & 38.4 & 40.9 & 44.6 & 36.8 & 19.9 & 23.3 & 53.7 & 21.5 & 32.5 & 37.3 & 32.9 & 41.1 & 35.6\\
VideollmOnline & 64 & 53.0 & 20.8 & 45.0 & 22.6 & 48.0 & 27.0 & 28.1 & 50.7 & 22.5 & 38.3 & 71.5 & 47.4 & 20.7 & 37.2\\
\bottomrule
\end{tabular}
}
\caption{Evaluation on Sub-tasks about Planning.}
\label{tab:table8_3}
\end{table*}

In this supplementary material, we provide additional details and analyses to complement the main paper. Section \ref{sec:sec_6} presents a comprehensive comparison of our proposed StreamEQA with existing video understanding benchmarks, along with extended data statistics. Section \ref{sec:sec_8} describes our experiments, including specific experimental implementations and more quantitative results on the performance of the models. Section \ref{sec:sec_7} dives deeper into the construction process of StreamEQA, specifically focusing on the information extraction from meta annotations. Finally, Section \ref{sec:sec_9} shows more data examples of each sub-task.


\section{Comparison with Existing Benchmarks}
\label{sec:sec_6}

As shown in Table~\ref{tab:table6}, StreamEQA sets itself apart from existing video understanding benchmarks by integrating both streaming and embodied task characteristics. Unlike StreamingBench~\cite{Lin2024StreamingBenchAT}, which only focuses on streaming tasks, or EgoThink~\cite{cheng2024egothink}, which focuses on embodied scenarios, StreamEQA uniquely combines both aspects, providing a more comprehensive challenge for video-LLMs. With an average video length of 67.5 seconds, 42 subtasks categories and 21K QA pairs, it surpasses other benchmarks in terms of task variety and dataset scale, offering a more rigorous and holistic evaluation of model generalization for real-world embodied agent application.

\section{More Experiments Results}
\label{sec:sec_8}

\subsection{Evaluation Protocol}
This section supplements the settings from Section~\ref{sec:4_1} to ensure full reproducibility of results reported in Section~\ref{sec:4_2} and a fair comparison across all models. 
 
Our evaluation is based on a zero-shot, single-round inference paradigm, executed on NVIDIA A6000 GPUs. As previously mentioned, video frames are sampled in fixed 64 frames (GPT-5 and Gemini-2.5-Pro are evaluated at 1 fps). To guarantee deterministic and reproducible outputs, we adhere to the official inference strategies of these MLLMs.
 
\begin{figure}[t]
  \centering
  \includegraphics[
    width=1.0\columnwidth,
    trim=8.2cm 6.0cm 8.2cm 6.0cm,
    clip
    ]{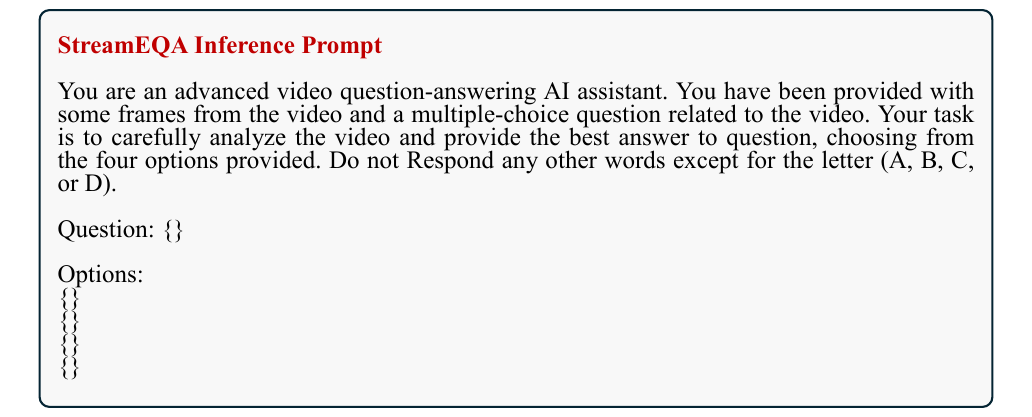}  
  \caption{Prompt of StreamEQA Inference}
  \label{fig:fig8_1}
\end{figure}

Furthermore, for multiple-choice questions, the prompt illustrated in Figure~\ref{fig:fig8_1} instructs the model to return its answer selected from A, B, C, or D.

\subsection{Sub-tasks Evaluation}
This section provides detailed results for all three embodied levels: Perception, Interaction, and Planning, broken down by task capabilities across different levels. The results are presented in Table~\ref{tab:table8_1}, Table~\ref{tab:table8_2}, and Table~\ref{tab:table8_3}.

\paragraph{Tasks of Perception.} As illustrated in Table~\ref{tab:table8_1}, most offline video-MLLMs perform relatively better in tasks of perception. Gemini-2.5-Pro and GPT-5 achieve the highest average accuracy, followed by EgoGPT. We can observe that Gemini-2.5-Pro performs better in ARB, RRB, ARR, RRR, and SUR, which contribute the most to the overall accuracy, while GPT-5 has the best performance in ARF, RRF, and SUF, indicating that the proprietary model is good at predicting objects and actions that may appear in the future. Notably, in addition to EgoGPT, some other offline open-sourced MLLMs outperform in some specific sub-tasks, such as MiniCPM-V and Qwen3VL, and their average accuracy is similar to proprietary models. This may reveal that the challenges of such tasks are limited. 

\paragraph{Tasks of Interaction.} Compared to perception tasks, all MLLMs face significant challenges in interaction tasks, with average scores below 61\% (Table~\ref{tab:table8_2}). In the settings, EgoGPT considerably outperforms open-source offline video-MLLMs. For instance, in the Episodic Memory task, EgoGPT surpasses Qwen3VL, which serves as the baseline for video understanding, by approximately 31\% (78.3\% vs. 47.6\%). Furthermore, EgoGPT also far exceeds GPT-5 in certain realtime and forward tasks, such as OER (73.0\% vs. 49.6\%), ARC (84.3\% vs. 61.2\%), ERD (49.5\% vs. 39.6\%), and ARP (85.7\% vs. 84.9\%). This indicates that explicitly designed embodied video-MLLMs for embodied scenarios hold significant advantages in interaction tasks.

\paragraph{Tasks of Planning.} All offline Video-MLLMs exhibit relatively low performance in tasks of planning, facing challenges in embodied reasoning (Table~\ref{tab:table8_3}). However, it is noteworthy that online video-MLLMs, such as Dispider and FlashVStream, have shown certain advantages compared with most offline Video-MLLMs, especially in forward tasks. For example, Dispider outperforms Qwen3VL by approximately 11\% in Steps Merging and 16\% in Moving-up Steps. Ultimately, this granular analysis confirms the primary bottleneck: the key challenge facing video-MLLM practical applications is embodied scenario analysis and temporal reasoning capabilities.

\section{Data Construction Details}
\label{sec:sec_7}
This section provides a more detailed description of the data construction process to ensure reproducibility, and offer deeper insight into how StreamEQA is systematically synthesized.

\subsection{Extraction of Meta Information}
Accurate meta informations is essential for QA construction. We extract such information from the meta annotations sourced from the HD-EPIC dataset~\cite{Perrett2025HDEPICAH}, which provide dense narrations with precise timestamps, event-level time ranges, eye-gaze priming, and spatial trajectories of objects. 
Specifically, we determine the temporal boundaries of each event (comprising multiple actions) within the video based on event time ranges and eye-gaze priming. The spatial relationships of objects are derived from the objects tracks, while meta informations of interaction and planning are inferred from the dense narrations. We will detail each key step of this extraction process in the following.

\paragraph{Object Spatial Relationship. }
The spatial relationship between two objects is computed directly from their bounding boxes, obtained as top-left and bottom-right coordinates from the objects’ spatial tracks. For each video frame, all co-occurring object pairs are evaluated. If two bounding boxes are identical or one is fully enclosed by the other, the relationship is assigned as ``same place''. Otherwise, we calculate the center point of each box (via the average of its diagonal coordinates) and compare their horizontal and vertical offsets to derive directional relations such as ``above'', ``below'', ``left'', or ``right''.
When both axes indicate displacement, composite relations (e.g., ``below-left'') are resolved using a predefined mapping to the closest semantically meaningful descriptor. Finally, unreasonable or visually implausible relations are filtered out using cues from narrative annotations, ensuring that the resulting spatial labels remain consistent with the actual scene dynamics.

\paragraph{Object Semantic Relationship. }
\begin{figure}[t]
  \centering
  \includegraphics[
    width=1.0\columnwidth,
    trim=8.2cm 0.2cm 8.2cm 0.2cm,
    clip
    ]{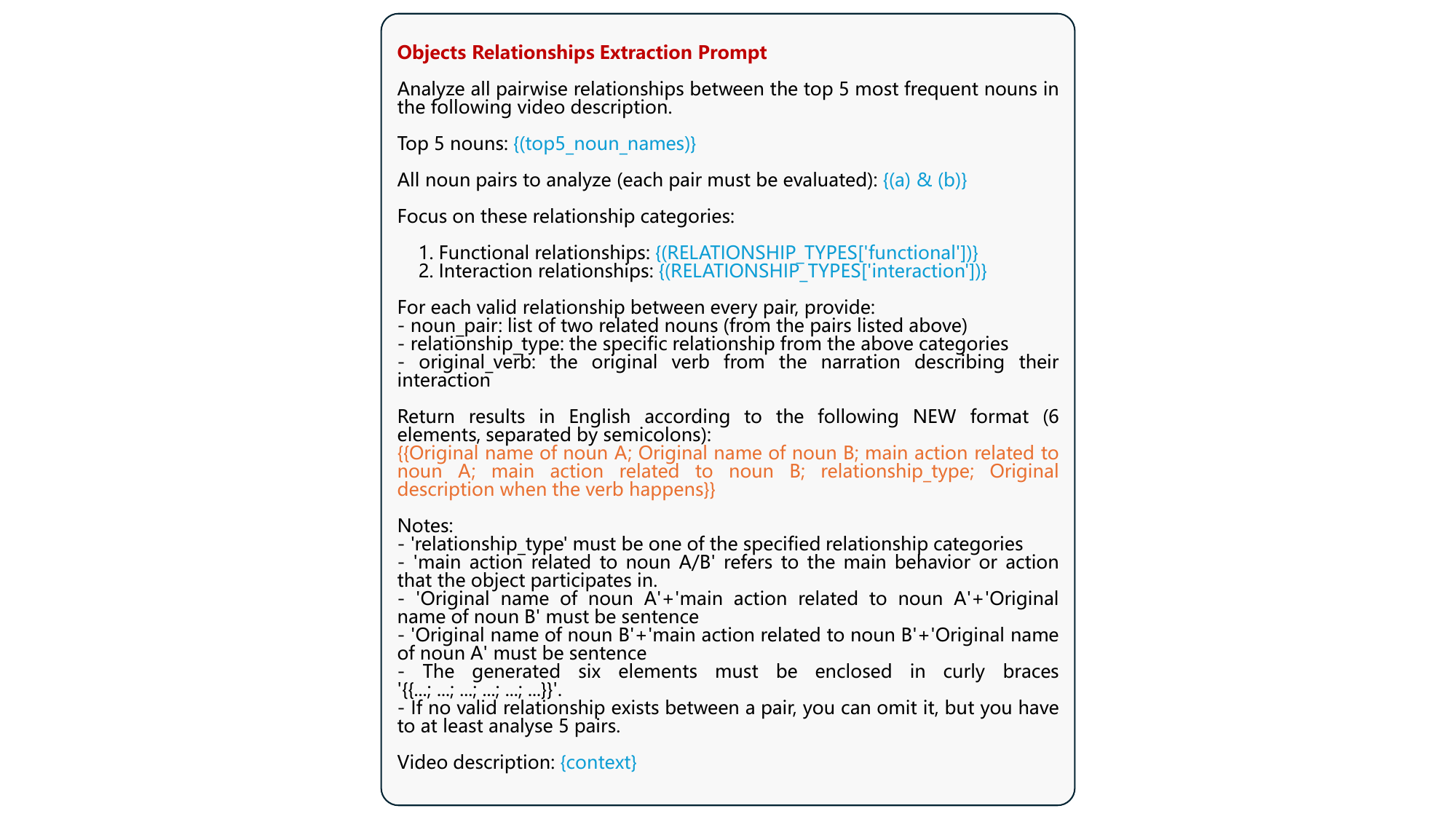}  
  \caption{Prompt of Object Semantic Relationship Extraction}
  \label{fig:fig7_1_1}
\end{figure}

We leverage GPT-5 to extract semantic relationships between objects from the meta annotations, with the prompt shown in Figure~\ref{fig:fig7_1_1}. Specifically, for each event, we first identify the five most frequent nouns from its dense narrations. We then analyze all pairwise combinations and consider two categories of relationships: functional relationships (e.g., ``contains'') and interaction relationships (e.g., ``assists''). For each object pair, GPT-5 is instructed to output a hexatuple consisting of the two object names, the actions occurring between them, and the corresponding relationship types.

\paragraph{Actions Motivation. }
\begin{figure}[t]
  \centering
  \includegraphics[
    width=1.0\columnwidth,
    trim=8.2cm 1.0cm 8.2cm 1.0cm,
    clip
    ]{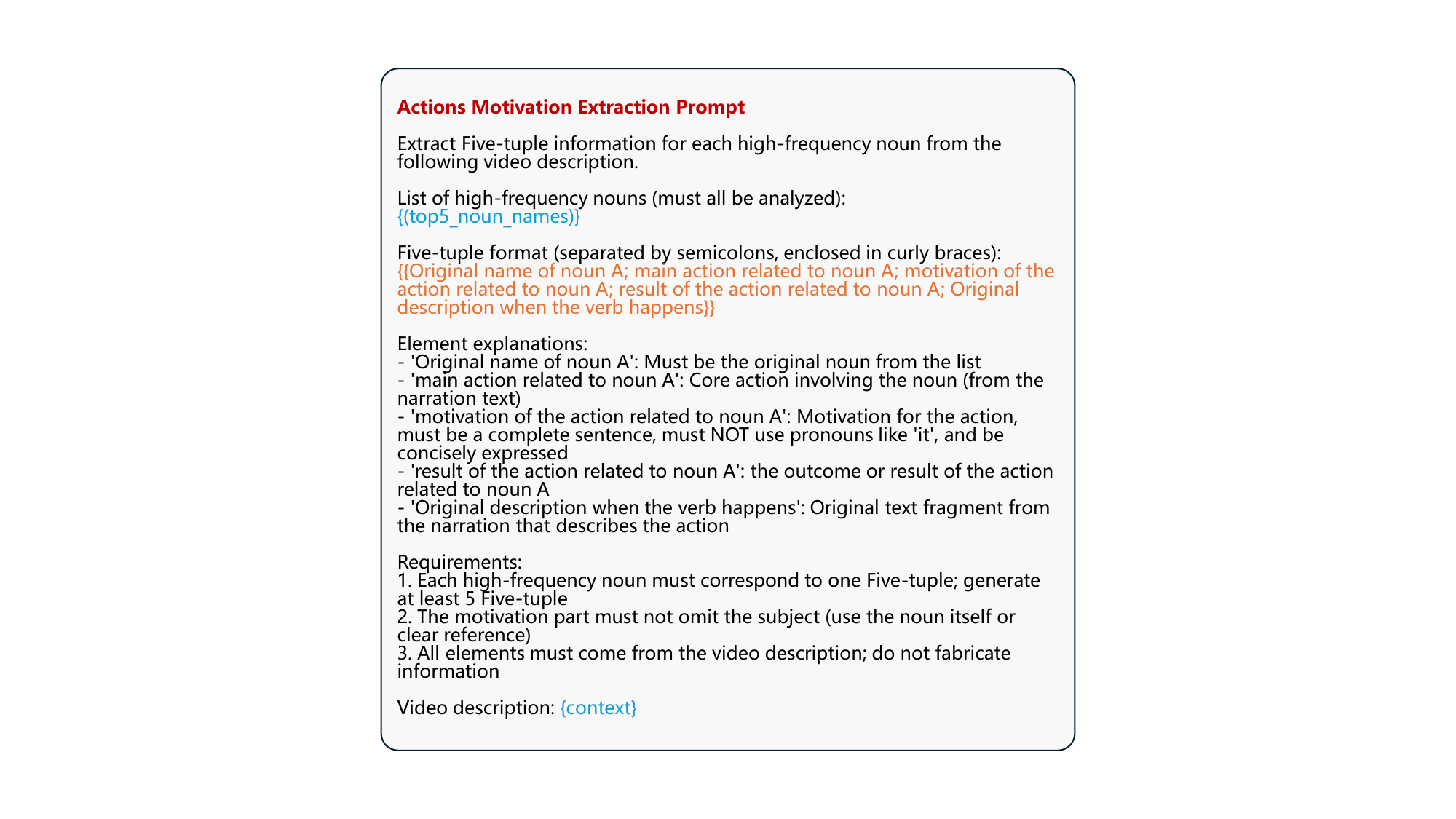}  
  \caption{Prompt of Actions Motivation Extraction}
  \label{fig:fig7_1_2}
\end{figure}

Figure~\ref{fig:fig7_1_2} illustrates the prompt for actions motivation extraction. The prompt is provided with the top 5 objects with the highest occurrence frequency within the event, and requires the model to extract the actions related to these objects, as well as analyze the motivations and outcomes of such actions.

\paragraph{Procedure Optimization. }
\begin{figure}[t]
  \centering
  \includegraphics[
    width=1.0\columnwidth,
    trim=8.2cm 0.1cm 8.2cm 0.1cm,
    clip
    ]{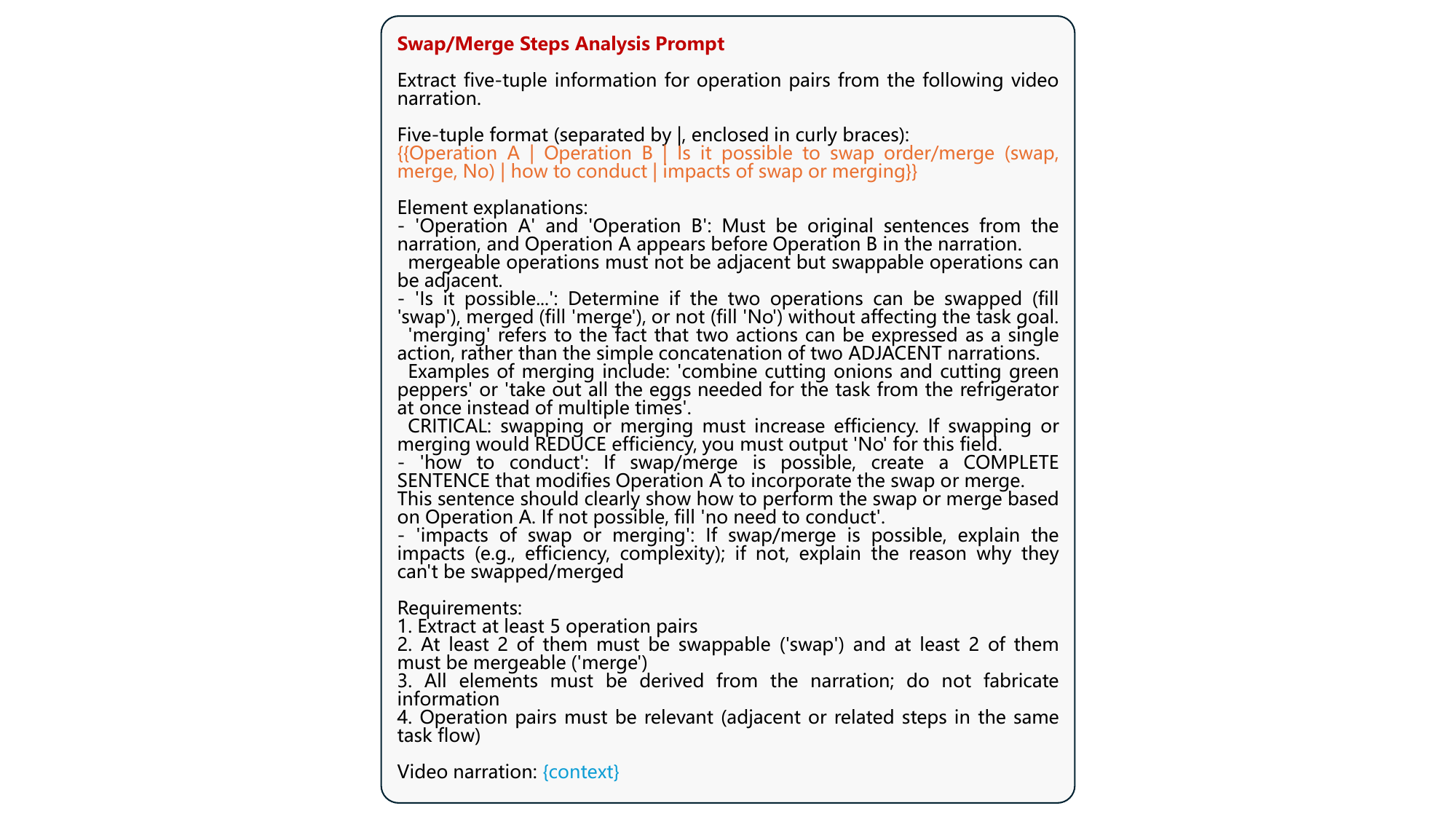}  
  \caption{Prompt of Swap/Merge Steps Analysis}
  \label{fig:fig7_1_3}
\end{figure}

\begin{figure}[t]
  \centering
  \includegraphics[
    width=1.0\columnwidth,
    trim=8.2cm 0.1cm 8.2cm 0.1cm,
    clip
    ]{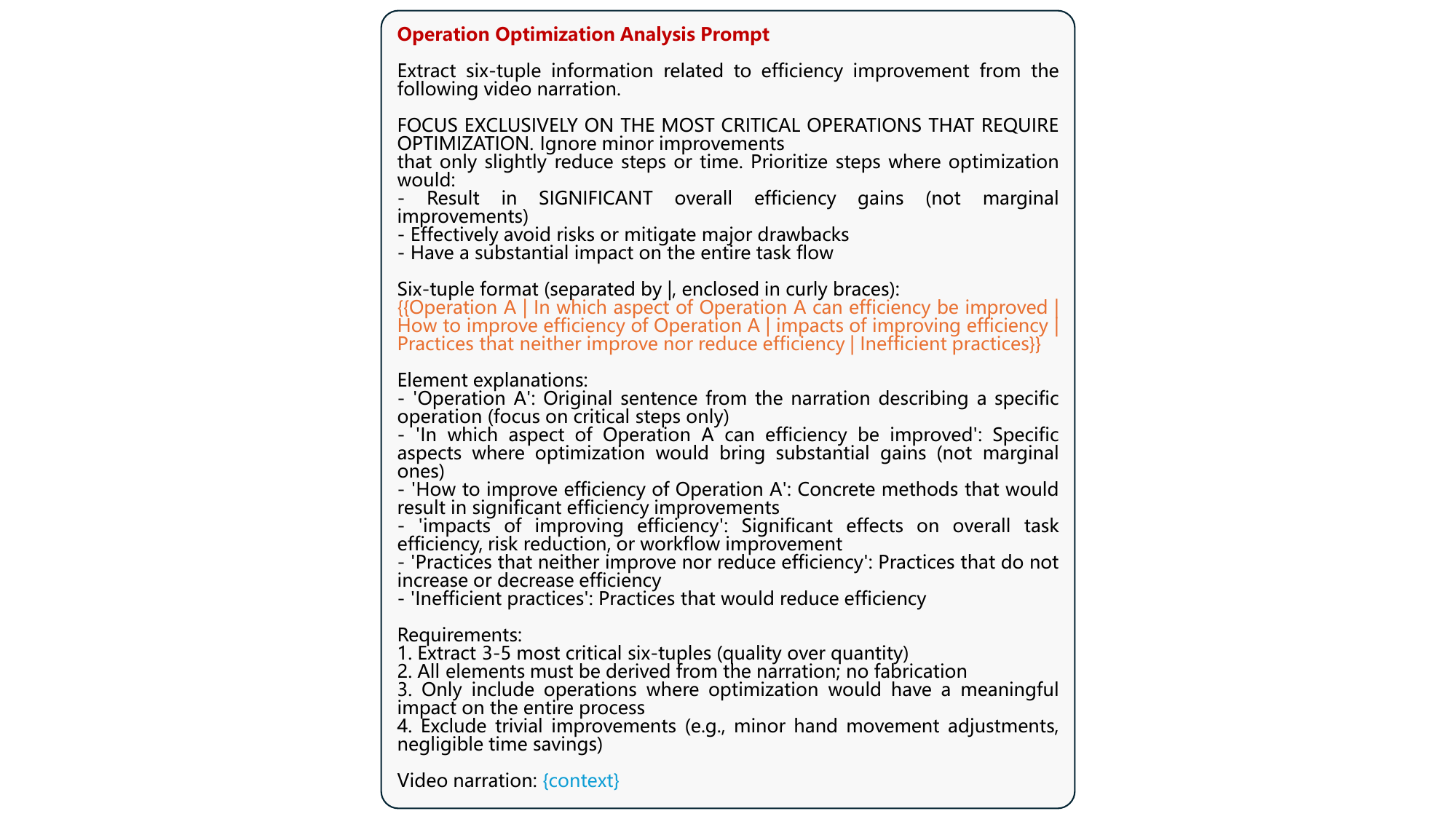}  
  \caption{Prompt of Operation Optimization Analysis}
  \label{fig:fig7_1_4}
\end{figure}

The Procedure Optimization Prompt comprises two components. As illustrated in Figure~\ref{fig:fig7_1_3}, the swap/merge steps analysis prompt instructs GPT-5 to analyze whether two operations described in the context can be merged or swapped to enhance efficiency. Additionally, we have designed a second prompt for single-operation optimization (Figure~\ref{fig:fig7_1_4}). For a given event, the prompt is designed to identify inefficient operations within the event and propose an optimized procedure for achieving the target.

\paragraph{Plan Reflection and Adjustment. }
\begin{figure}[t]
  \centering
  \includegraphics[
    width=1.0\columnwidth,
    trim=8.2cm 0.0cm 8.2cm 0.05cm,
    clip
    ]{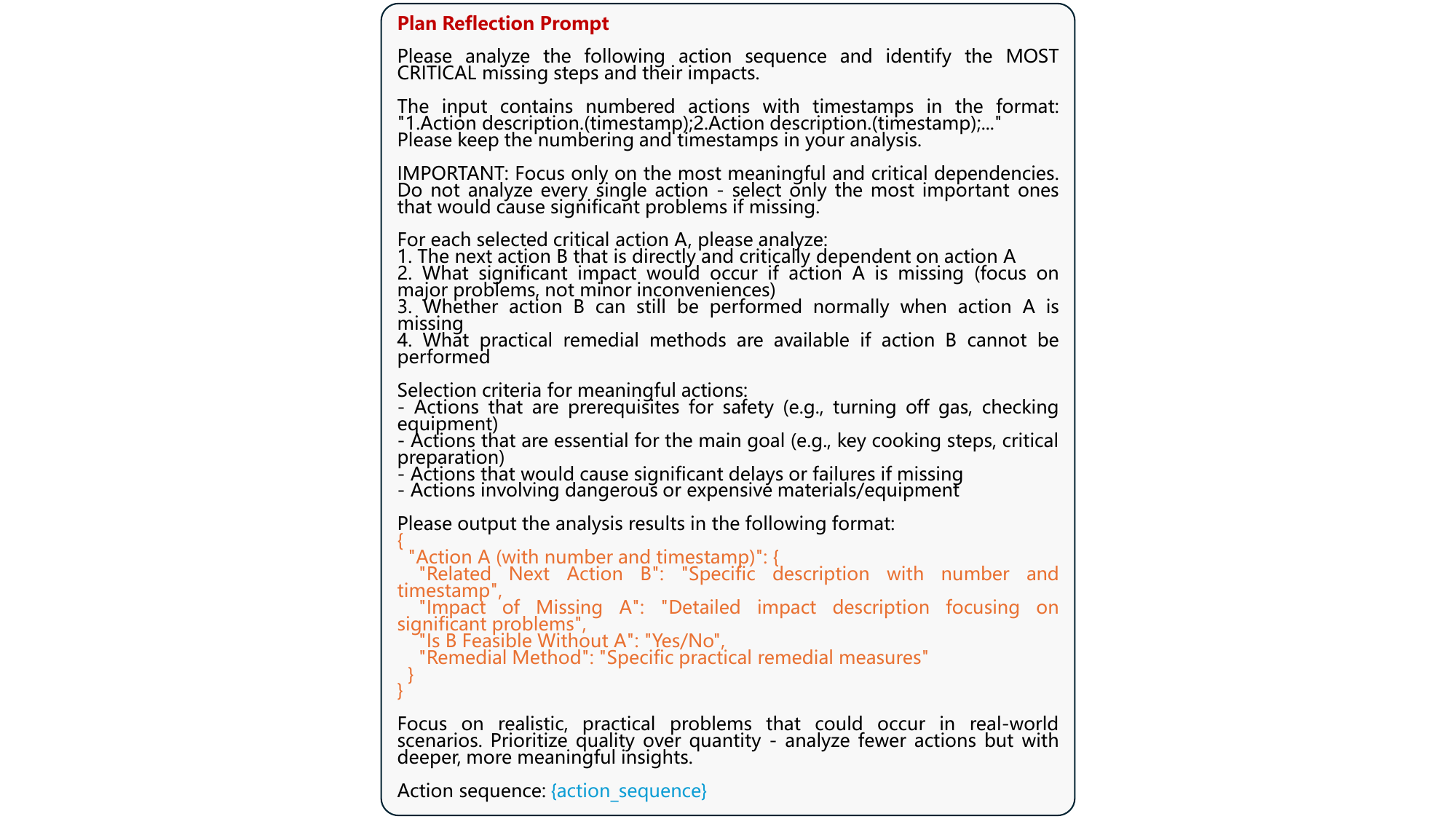}  
  \caption{Prompt of Plan Reflection}
  \label{fig:fig7_1_5}
\end{figure}

\begin{figure}[t]
  \centering
  \includegraphics[
    width=1.0\columnwidth,
    trim=8.2cm 0.7cm 8.2cm 0.7cm,
    clip
    ]{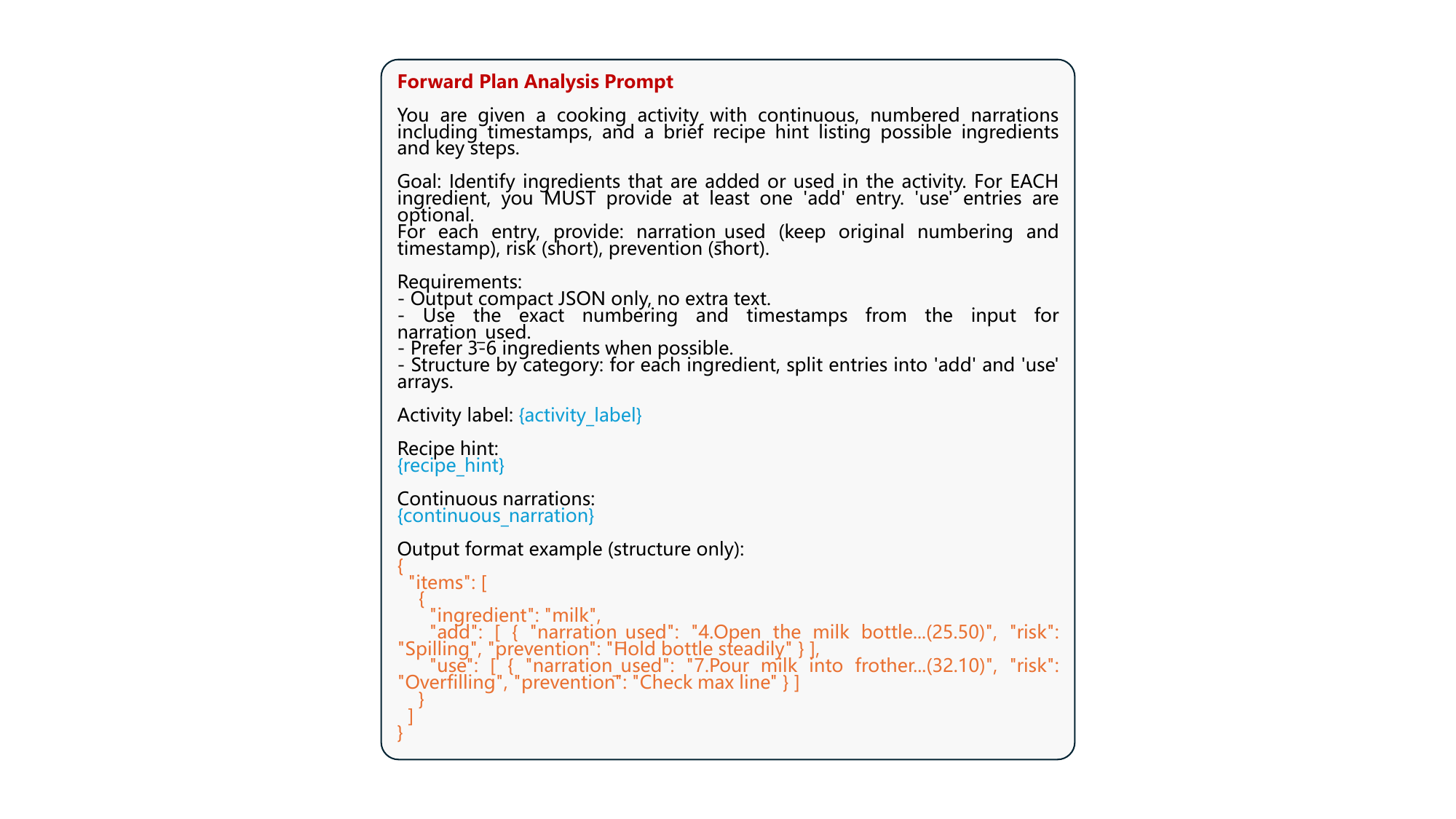}  
  \caption{Prompt of Forward Plan Analysis}
  \label{fig:fig7_1_6}
\end{figure}

To generate QA pairs for evaluating models' capabilities to reflect on past procedures and adjust future plans accordingly, we have designed two prompts. As illustrated in Figure~\ref{fig:fig7_1_5}, the Plan Reflection Prompt analyzes an action sequence to identify the most critical actions, assessing their significant impacts if missing, the feasibility of subsequent actions, and practical remedial measures. In addition, we have designed the Forward Plan Analysis Prompt (Figure~\ref{fig:fig7_1_6}), which is tailored to cooking activities to identify ingredients incorporated in the cooking process. For each ingredient, the prompt requires GPT-5 to analyze three key aspects in a structured JSON format: when and how to adjust the ingredient usage, the risks associated with relevant operations, and corresponding preventive measures.

\subsection{QA Refinement}
\begin{figure}[t]
  \centering
  \includegraphics[
    width=1.0\columnwidth,
    trim=8.2cm 3.5cm 8.2cm 3.5cm,
    clip
    ]{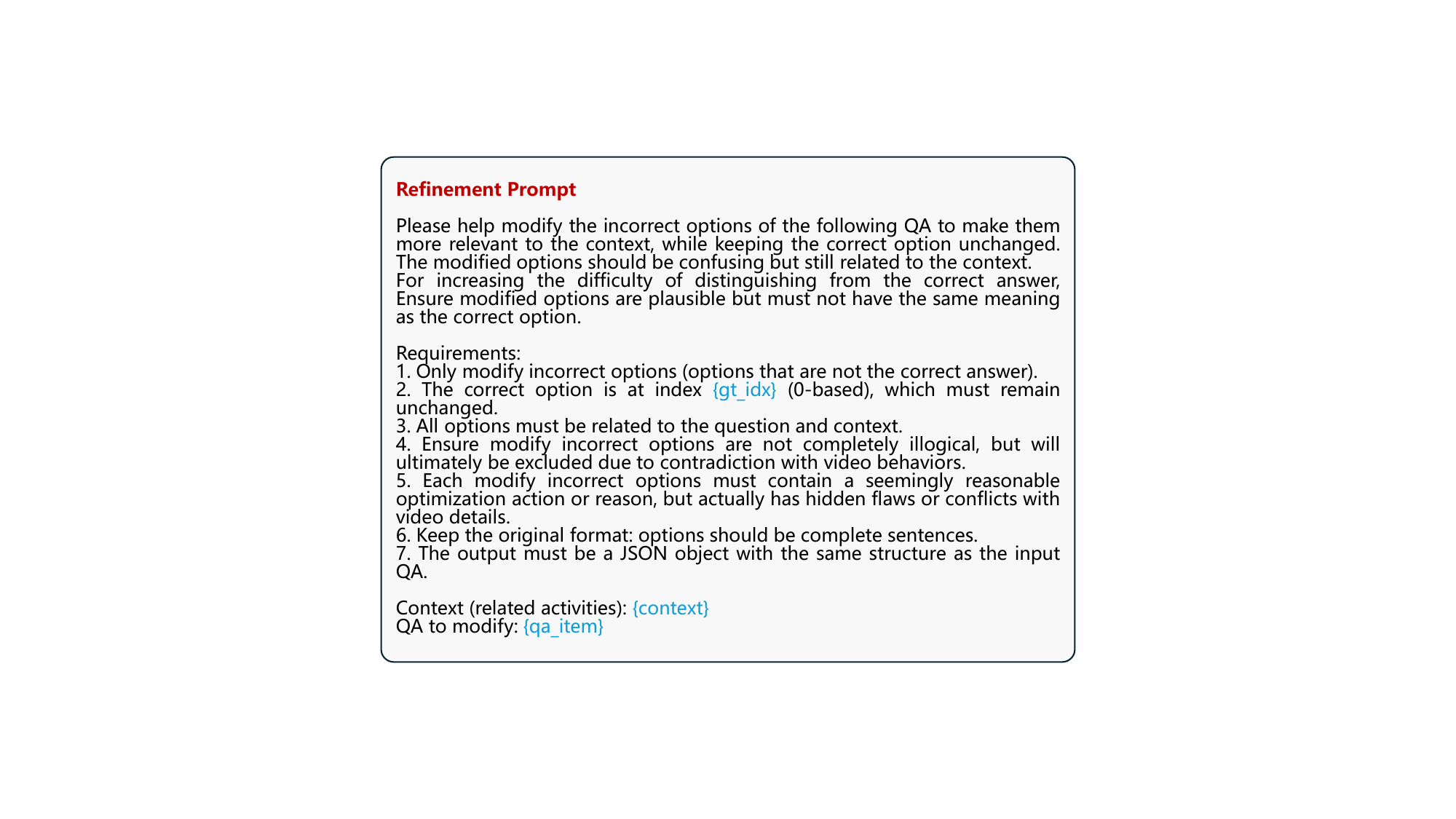}  
  \caption{Prompt of QA Refinement}
  \label{fig:fig7_2}
\end{figure}

To ensure StreamEQA can effectively evaluate capabilities about video understanding and reasoning, QA Refinement is introduced to address the limitations of originally generated distractors (incorrect options), such as contextual irrelevance or weak, misleading nature.
When conducting QA refinement, several key aspects need to be highlighted. Firstly, only distractors should be modified, while the correct option (gt\_idx) remains unchanged. Furthermore, modified incorrect options must be contextually related and plausible but ultimately excluded due to contradictions with video context, and distractors should not share the same meaning as the correct option. In addition to QA pairs (qa\_item) that need to be refined, we provide context to supplement the details and relationships of relevant activities, as shown in Figure~\ref{fig:fig7_2}.

\section{More Data Examples}
\label{sec:sec_9}

\begin{figure*}[t]
  \centering
  \includegraphics[
    width=0.85\textwidth,  
    trim=0.3cm 1.7cm 0.3cm 1.7cm,  
    clip
  ]{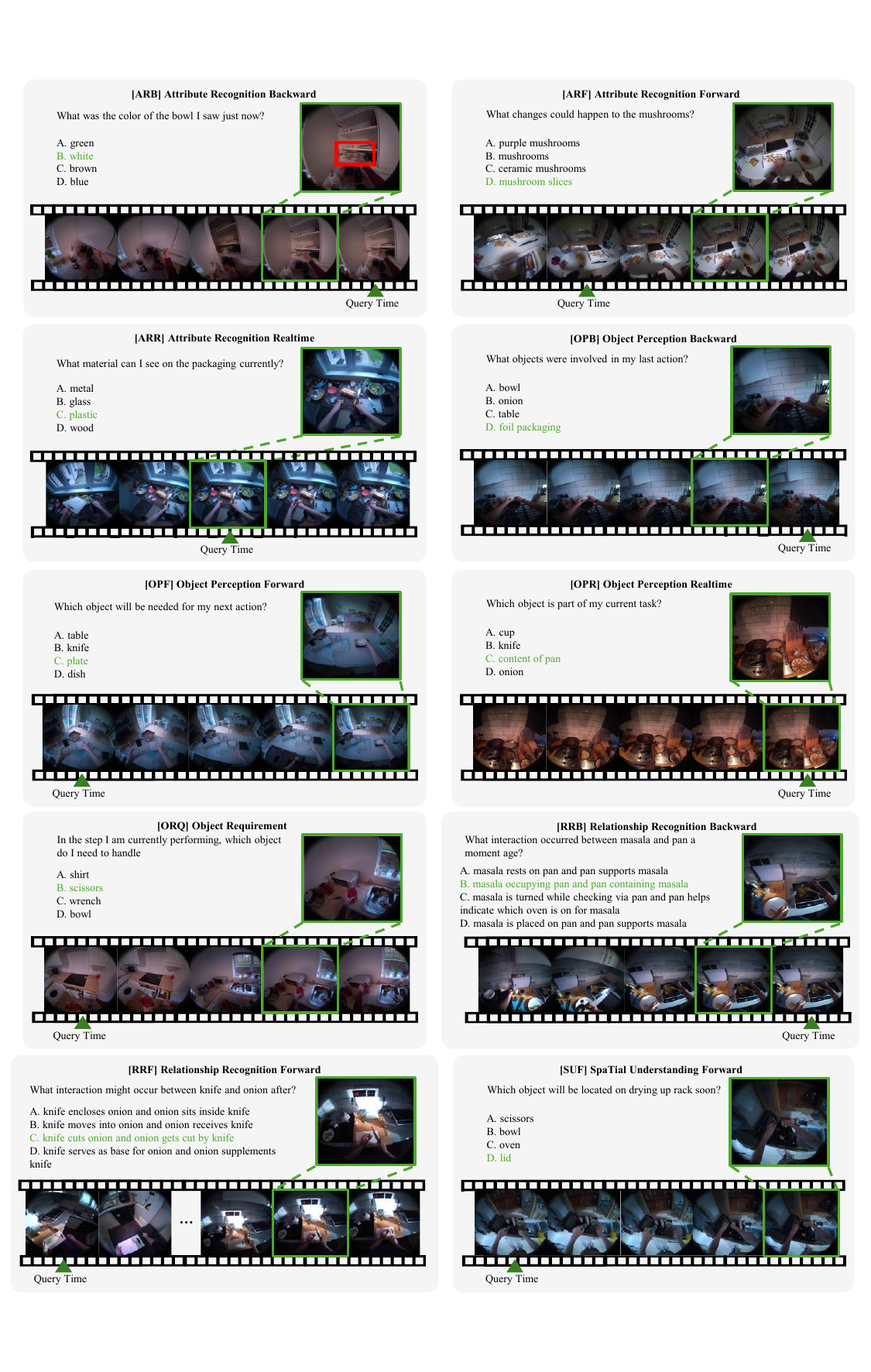}
  \caption{Data examples for Attribute Recognition Backward, Attribute Recognition Forward, Attribute Recognition Realtime, Object Perception Backward, Object Perception Forward, Object Perception Realtime, Object Requirement, Relationship Recognition Backward, Relationship Recognition Forward, SpaTial Understanding Forward tasks.}
  \label{fig:fig9_1}
\end{figure*}

\begin{figure*}[t]
  \centering
  \includegraphics[
    width=0.9\textwidth,  
    trim=0.3cm 3.5cm 0.3cm 3.5cm,  
    clip
  ]{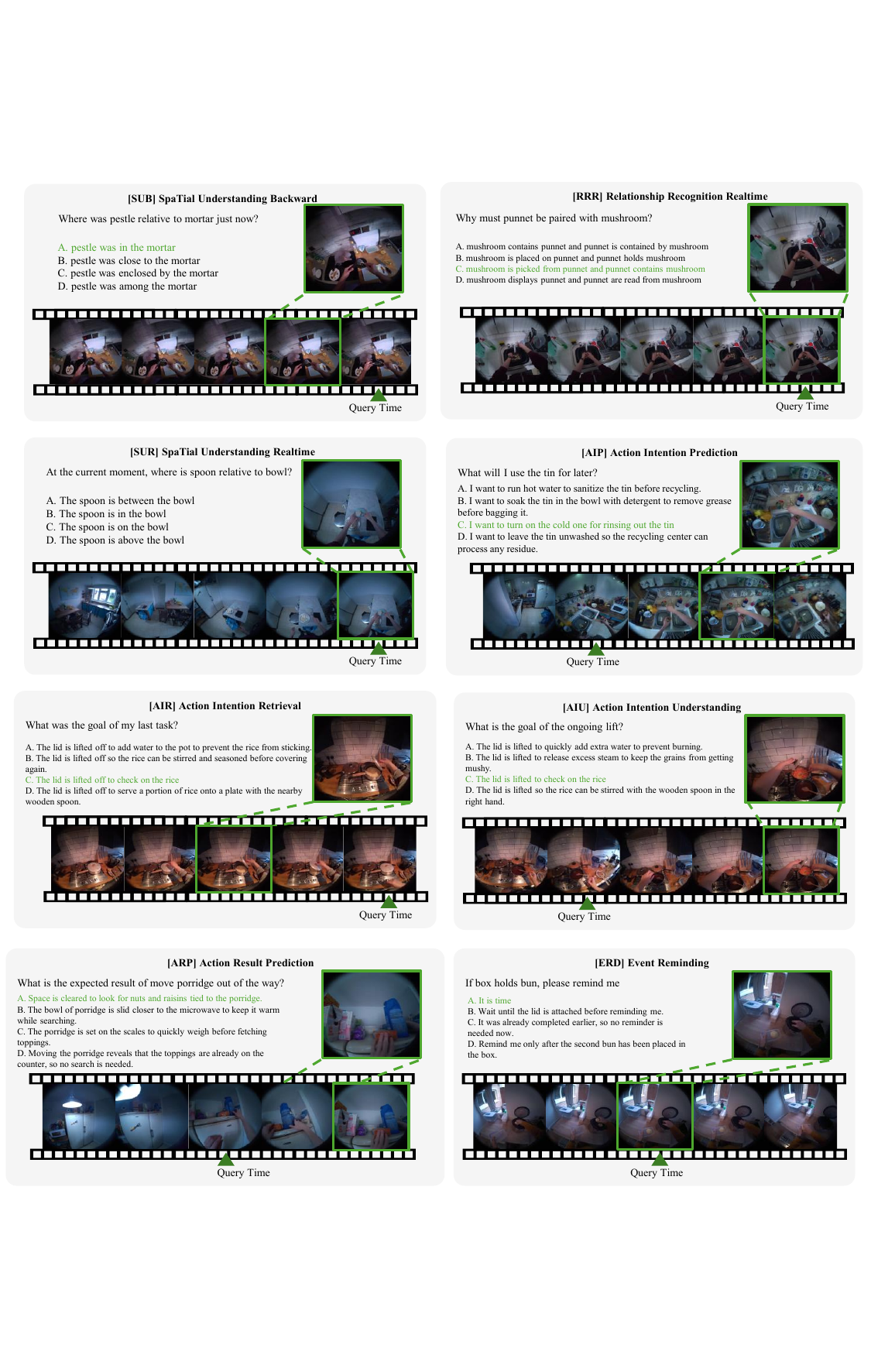}
  \caption{Data examples for SpaTial Understanding Backward, Relationship Recognition Realtime, SpaTial Understanding Realtime, Action Intention Prediction, Action Intention Retrieval, Action Intention Understanding, Action Result Prediction, Event Reminding tasks.}
  \label{fig:fig9_2}
\end{figure*}

\begin{figure*}[t]
  \centering
  \includegraphics[
    width=0.9\textwidth,  
    trim=0.3cm 3.5cm 0.3cm 3.5cm,  
    clip
  ]{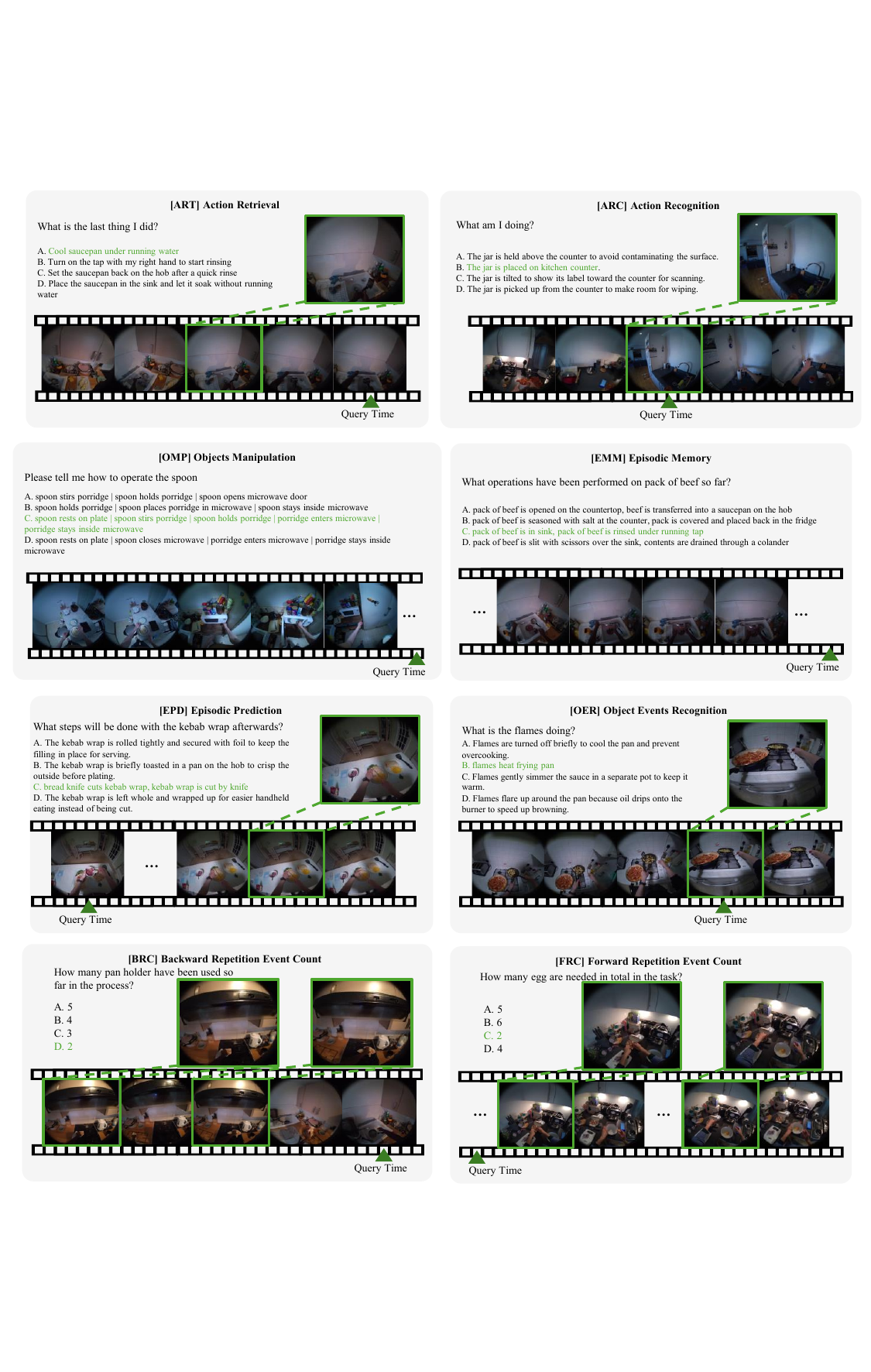}
  \caption{Data examples for Action Retrieval, Action Recognition, Objects Manipulation, Episodic Memory, Episodic Prediction, Object Events Recognition, Backward Repetition Event Count, Forward Repetition Event Count tasks.}
  \label{fig:fig9_3}
\end{figure*}

\begin{figure*}[t]
  \centering
  \includegraphics[
    width=0.9\textwidth,  
    trim=0.3cm 3.5cm 0.3cm 3.5cm,  
    clip
  ]{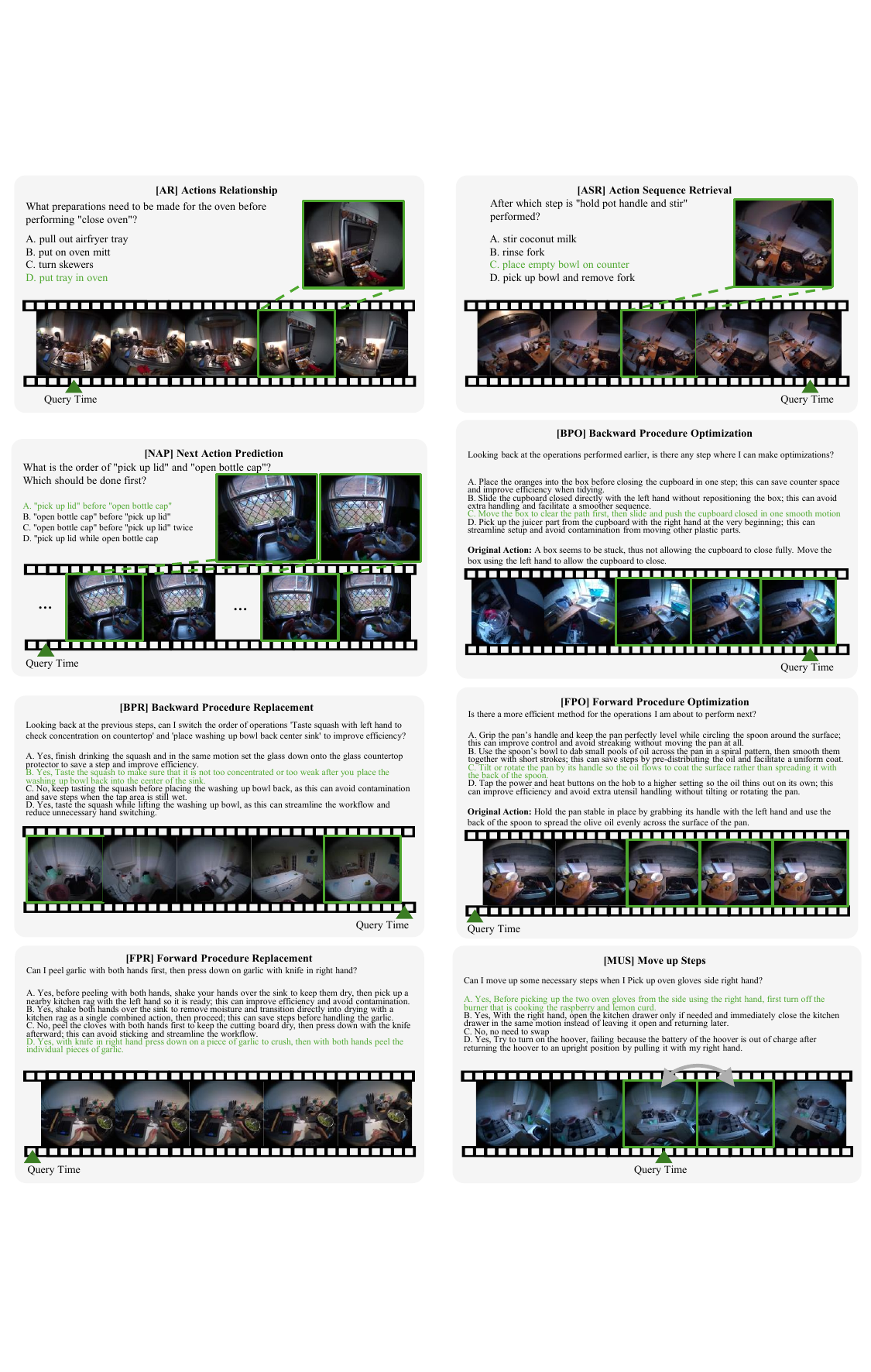}
  \caption{Data examples for Actions Relationship, Action Sequence Retrieval, Next Action Prediction, Backward Procedure Optimization, Backward Procedure Replacement, Forward Procedure Optimization, Forward Procedure Replacement, Move up Steps tasks.}
  \label{fig:fig9_4}
\end{figure*}

\begin{figure*}[t]
  \centering
  \includegraphics[
    width=0.9\textwidth,  
    trim=0.3cm 3.5cm 0.3cm 3.3cm,  
    clip
  ]{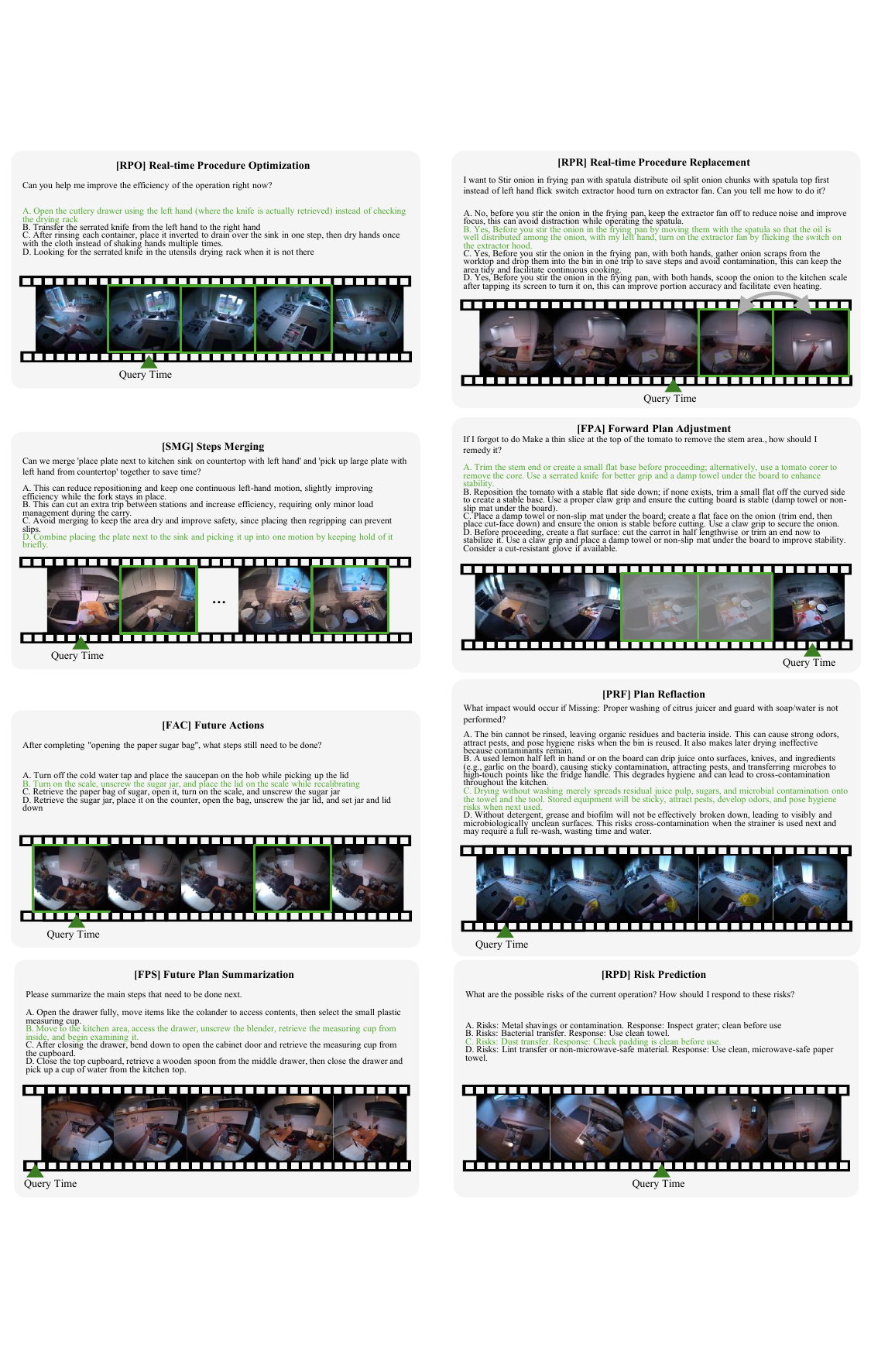}
  \caption{Data examples for Real-time Procedure Optimization, Real-time Procedure Replacement, Steps Merging, Forward Plan Adjustment, Future Actions, Plan Reflection, Future Plan Summarization, Risk Prediction tasks.}
  \label{fig:fig9_5}
\end{figure*}

As shown in Figure~\ref{fig:fig9_1} to Figure~\ref{fig:fig9_5}, we provide more examples extracted from our benchmark. We try to cover different tasks to offer a holistic overview of StreamEQA.

\end{document}